\documentclass[11pt, a4paper]{article}

\title{\Large A Causal Perspective on Loan Pricing\\
       \large Investigating the Impacts of Selection Bias on Identifying Bid-Response Functions}

\date{September, 2023}

\author{
        Christopher Bockel-Rickermann
        \thanks{Faculty of Economics and Business, KU Leuven}
        \and
        Sam Verboven
        \thanks{Faculteit Economische en Sociale Wetenschappen en Solvay Business School, Vrije Universiteit Brussel}
        \and
        Tim Verdonck
        \thanks{Department of Mathematics, University of Antwerp - imec}
        \thanks{Department of Mathematics, KU Leuven}
        \and
        Wouter Verbeke\footnotemark[1]
        }

\usepackage{amssymb}
\usepackage{amsthm}
\usepackage{amsthm}
\usepackage{amscd}
\usepackage{amsfonts}
\usepackage{bm}
\usepackage{amsmath}
\usepackage{verbatim} 
\usepackage{courier}
\usepackage{arydshln}
\usepackage{adjustbox}
\usepackage{multirow}
\usepackage{longtable}
\usepackage[T1]{fontenc}
\usepackage{comment}
\usepackage{xcolor}
\usepackage{soul}
\usepackage{caption}
\usepackage{subcaption}
\usepackage{enumitem}
\usepackage{tikz}
\usepackage{setspace}
\usepackage{hyperref}
\usepackage[sort,authoryear,compress]{natbib}
\usepackage[left=72pt, right=72pt, right=72pt, top=72pt]{geometry}

\hypersetup{
    colorlinks = true,
    citecolor  = blue,
    linkcolor  = blue,
}

\newcommand{\tikzxmark}{
\tikz[scale=0.23] {
    \draw[line width=0.7,line cap=round] (0,0) to [bend left=6] (1,1);
    \draw[line width=0.7,line cap=round] (0.2,0.95) to [bend right=3] (0.8,0.05);
}}
\newcommand{\tikzcmark}{
\tikz[scale=0.23] {
    \draw[line width=0.7,line cap=round] (0.25,0) to [bend left=10] (1,1);
    \draw[line width=0.8,line cap=round] (0,0.35) to [bend right=1] (0.23,0);
}}

\usepackage{titlesec}
\titleformat{\section}{\bfseries\fontsize{18pt}{18pt}\selectfont}{\thesection}{0.618em}{}
\titleformat{\subsection}{\bfseries\fontsize{15pt}{15pt}\selectfont}{\thesubsection}{0.618em}{}

\newtheorem{assumption}{Assumption}

\begin{document}

\setcounter{page}{1} \renewcommand{\thepage}{\arabic{page}}

\maketitle

\begin{abstract}
    \noindent In lending, where prices are specific to both customers and products, having a well-functioning personalized pricing policy in place is essential to effective business making. Typically, such a policy must be derived from observational data, which introduces several challenges. While the problem of ``endogeneity'' is prominently studied in the established pricing literature, the problem of selection bias (or, more precisely, bid selection bias) is not. We take a step towards understanding the effects of selection bias by posing pricing as a problem of causal inference. Specifically, we consider the reaction of a customer to price a treatment effect. In our experiments, we simulate varying levels of selection bias on a semi-synthetic dataset on mortgage loan applications in Belgium. We investigate the potential of parametric and nonparametric methods for the identification of individual bid-response functions. Our results illustrate how conventional methods such as logistic regression and neural networks suffer adversely from selection bias. In contrast, we implement state-of-the-art methods from causal machine learning and show their capability to overcome selection bias in pricing data.\\[0.25cm]
\end{abstract}
	
\section{Introduction\label{sec:Intro}}

\noindent Pricing loans is a challenging task for banks. A pricing policy must take into account and balance several risks to be profitable. Prices must be both high enough to account for operating costs and losses and low enough to match price sensitivities of customers and to be competitive in the overall market. If a price is too high, customers may reject it and seek a loan at a competitor. However, if a price is too low, the bank may not turn the loan into a profit and possibly suffer losses. 

As customers and their preferences are heterogeneous, an optimal price in terms of revenue or profit maximization is a personalized one \citep[]{Varian.1989}. In reality, many pricing policies are lacking systematic and data-driven personalization. Instead, prices are typically derived from rigid pricing grids that segment customers into large groups. Individual components, such as a discount, are typically applied at the discretion of a bank clerk incentivized by corporate policies \citep[]{Phillips.2015}.

Developing a personalized pricing policy to optimize loan prices is challenging since a bank must know the individual preferences and characteristics of a customer. Typically, these characteristics, such as price sensitivity and willingness to pay, are hidden. Banks must assume or approximate those characteristics to build well-functioning models \citet{Phillips.2021}. These approximations prove complicated as they rely on observational data of past loan negotiations, introducing several challenges: while the problems of endogeneity and  hidden confoundedness are generally prominently studied in the established pricing literature \citep[]{Berry.1994, Villas.1999},  selection bias \citep[]{Heckman.1990} is not. Selection bias (or ``bid selection bias'', as in the context of this work) persists in observational pricing data due to bias in the established pricing policy. Certain customer groups might receive preferential treatment in accordance with an established pricing grid or regulatory requirements \citep[]{Jain.2016}, bank clerks might suffer from unconscious bias in handling customers and applying discounts, and customer groups might self-select in choosing which financial products to purchase.

Our work aims to study the impact of selection bias on learning personalized pricing models from observational data and builds on identifying methodologies that overcome potential adverse effects of selection bias. In order to do so, we propose to frame loan pricing as a problem of causal inference and look at pricing as a case of treatment and effect. The offered loan price, the ``bid'', is considered to be a continuously-valued treatment, whereas the treatment effect is the probability of a customer accepting this bid, i.e., the ``bid response''. A model of the individual treatment effect, that is, the individual bid response, can be used to optimize and personalize pricing. Applications include decision support for bank clerks, data-driven allocation of personalized prices, and improved customer segmentation in existing pricing policies.

To study the effects of bid selection bias, we work with a semi-synthetic dataset created in collaboration with a bank in Belgium. We apply a set of machine learning techniques, including novel methods from causal machine learning, to estimate bid-response curves and evaluate their robustness to varying levels of selection bias in the training data.

To the best of our knowledge, our study is the first of its kind and adds to the literature in the following ways:

\begin{itemize}[topsep=1pt, itemsep=1pt]
    \item we frame the pricing problem as a problem of causal inference;
    \item we showcase the issue of selection bias in observational pricing data and its impact on decision making;
    \item we evaluate the robustness of established and novel machine learning methods against bid selection bias in observational data.
\end{itemize}
The remainder of our work is organized as follows: in Section~\ref{sec:LitRev}, we provide an overview of the established literature on customized pricing. Section~\ref{sec:Method} defines pricing as a problem of causal inference and discusses reasons for selection bias, as well as methods for bid-response estimation. Section~\ref{sec:Experiments} describes our experimental evaluation, including the generation of our dataset. We conclude in Section~\ref{sec:Conclusion}, in which we reflect on results and discuss potential future research.

\section{Related Work\label{sec:LitRev}}

\noindent \textbf{Customized pricing:} Loan pricing is a case of customized pricing \citep[]{Phillips.2012, Phillips.2013}. Contrary to the pricing of  uniform goods \citep[see, e.g., ][]{Varian.2014}, in customized pricing, neither goods nor prices are homogeneous across customers \citep[]{Phillips.2021}. In loan pricing, customized products and prices can better meet customers' financing demands and offer banks a way to manage and mitigate risks. Customization of the product occurs by varying the duration, amount, and starting principal of the loan. Additionally, there typically is no price transparency, as customers cannot observe prices offered to other individuals. This changes the objective from setting a single best price for a customer group to setting the optimal price per individual. As a result, customized pricing is often considered an optimization problem \citep[]{Oliver.2014, Sundararajan.2011, Phillips.2012} that aims to optimize a certain business objective. We summarize the customized pricing process as such: after a customer has approached a seller, they request a price for a product, commonly referred to as a \textit{bid}. The seller will try to determine the bid that maximizes a certain objective, for example, revenue or profit generation per customer, which is facilitated by estimating the probability that a bid is accepted. The probability  of acceptance as a function of the height of the bid is referred to as the \textit{bid-response function}. It differs from the demand curve in microeconomics by representing a probability concerning a single individual instead of an aggregated demand over a customer group. Hereby, customized pricing is related to the wider field of market response modeling, to which an introduction can be found in, e.g., \citet{Hanssens.2003}. For the case of loans, we  describe an exemplary loan pricing and sales process in Section~\ref{sec:Method_Problem}.\\

\noindent \textbf{Endogeneity in pricing:} Traditionally, estimation of the bid response is approached with parametric and semiparametric methods, assuming a certain functional form of the bid response \citep[]{Phillips.2021, vanRyzin.2012}. This estimation is subject to various challenges in terms of, for example, the available information and infrastructure, model complexity, and legal and regulatory requirements \citep{Phillips.2012}. The most prominent challenge discussed in the established pricing literature is ``endogeneity'' in observational training data (or ``hidden confoundedness'' in the wider sense). Endogeneity is present if variables critical to explaining customer preferences or their actions are unobserved (or not available for modeling) \citep[]{Phillips.2012a}. Reasons for endogeneity are plentiful, stemming, for example, from interactions between customers and sales representatives that are not or insufficiently recorded, due to internal and external regulations. There is a wide body of literature on identifying and handling endogeneity, both in pricing and more generally in economics and statistics \citep[see, e.g., ][]{Heckman.1978, Newey.1987, Rivers.1988, Kuksov.2008}. De facto standard methods are instrumental variable regression \citep[e.g., ][]{Hausman.2012}, and BLP procedures named after the authors of \citet{Berry.1995}. More recently, machine learning has been proposed to finding bid-response models \citep[]{Lawrence.2003, Agrawal.2007, Arevalillo.2019, Guelman.2014}.\\

\noindent \textbf{Selection bias in pricing:} The impact of selection bias (cf. Section~\ref{sec:Method_SelecBias}) on pricing has received little attention in the literature. There is, however, an established body of research from other domains on the issue, such as medicine \citep[]{Berrevoets.2020}, economics \citep[]{Varian.2016}, market modeling \citep[]{Ferkingstad.2011}, and maintenance \citep[]{Vanderschueren.2022}. 

\section{Methodology\label{sec:Method}}

\noindent To showcase the impact of selection bias on customized pricing, we propose a methodology in four steps. In Section~\ref{sec:Method_Problem}, we describe the general loan pricing and sales process and frame it as a problem of causal inference. In Section~\ref{sec:Method_SelecBias}, we build on this framing and introduce selection bias in observational data and its relevance towards pricing. In Section~\ref{sec:Method_Assumptions}, we discuss the necessary assumptions for the causal identification of bid responses. Finally, in Section~\ref{sec:Method_Predict}, we present the methods that we apply to learn bid-response functions from observational data. Our experimental setup and results follow in Section~\ref{sec:Experiments}.

\subsection{Formalizing the loan pricing and sales process\label{sec:Method_Problem}}

\begin{figure}[ht]
    \centering
    \includegraphics[width=0.6\textwidth]{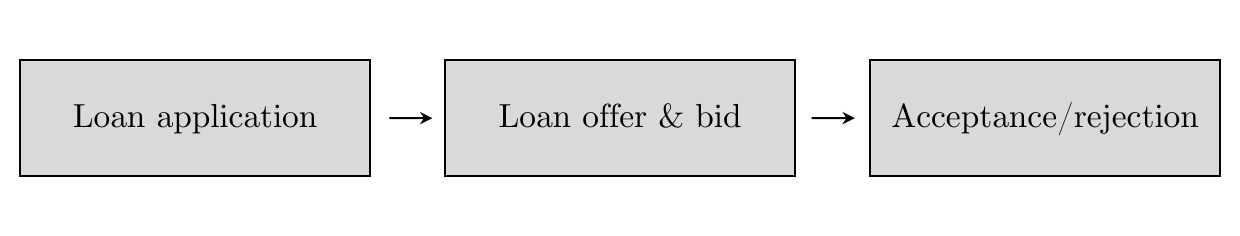}
    \caption{Simplified loan pricing process}
    \label{fig:loan_pricing_process}
\end{figure}

The general process of a customer taking out a loan is formalized as follows: 
\begin{itemize}
    \item first, a customer approaches a finite number of banks to negotiate a loan;
    \item second, a bank obtains information about the customer and evaluates whether and at what conditions it wants to grant the loan;
    \item third, each bank either rejects the customer or provides loan conditions, including a prospective price, i.e., the ``bid'';
    \item fourth and finally, the customer evaluates the offers received from the banks and decides whether to accept any of them, and if so, which. A round of negotiations might follow after the initial bid is placed. For the purpose of this study, we do not investigate any negotiations but focus on the first response of a customer to the first bid only.
\end{itemize}
For the purpose of our study, we simplify the process and look at it from the perspective of a single bank. The process reduces to three steps: a customer approaches the bank, the bank provides a loan offer including a bid for the price of the loan, and finally a customer decides on acceptance or refusal of the offer. The simplified process is depicted in Figure~\ref{fig:loan_pricing_process}.\\

\noindent \textbf{Loan price decomposition:} In practice, a loan price often consists of multiple parts, such as a base price and a customer-specific component. The base price serves to cover risks and costs to the bank and is determined via risk-based pricing. For risk-based pricing, banks cluster customers into a finite number of risk bands and determine a price per risk band that covers running costs of the bank, as well as expected losses \citep{Phillips.2013}. Risk models are typically proprietary, resulting in price dispersion in the market. The individual component to a loan price is given by a bank clerk during the sales process based on their information and personal assessment of the customer. Discounts might be driven by incentive schemes resulting from corporate strategy. The impact of such incentive schemes \citep{Garrow.2006, Phillips.2015} and the risk of principal-agent problems \citep[]{Berhold.1971,Ross.1973} is out of scope in this work. Additionally, we do not take into account the impacts of negotiations between customers and bank clerks that might occur.\\

\noindent \textbf{Loan pricing as a problem of causal inference:} We assume that a bank has records on their historical sales processes available, which are formalized as such: after a customer $i$ has decided about a specific loan offer, the result is summarized as a tuple $(\mathbf{x}_{i},b_{i,f},y_{i,f})$. All values are realizations of random variables $(\mathbf{X},B,Y)$. $\mathbf{X} \in \mathbb{R}^{d}$ describes the characteristics of the customer, as well as the specifications of the loan, excluding the bid, and information on competitor offers\footnote{We assume that all information relevant to the customer decision is known and in $\mathbf{X}$ to satisfy necessary assumptions (cf. Section~\ref{sec:Method_Assumptions}).}. $B \in \mathbb{R}$ is the bid made to the customer for the loan price, i.e., the credit rate in percent of interest per year. $Y \in \{0, 1\}$ is the decision to accept the offer and take up the loan or not. We assume $Y$ to be a Bernoulli random variable that is dependent on $P \in [0,1]$, the probability of accepting a loan offer; hence, $Y \sim Bernoulli(P)$.

In framing loan pricing as a problem of causal inference, we adopt the Neyman-Rubin potential outcome framework \citep[]{Rubin.2004, Rubin.2005} and assume that for every realization of a bid $b$ we find a potential outcome $P(b)$ of the probability of accepting the loan offer and a realization $Y(b) \in \{0,1\}$. All observed data and corresponding probabilities are referred to as ``factual'', including the factual bid $B_{f}$, the factual acceptance probability $P_{f} = P(B_{f})$, and the factual acceptance $Y_{f} \sim Bernoulli(P_{f})$. Any potential outcomes to unobserved combinations of a customer characteristic and a bid are referred to as ``counterfactual outcomes''.\\

\noindent \textbf{Bid response as a treatment effect:} For our study, we assume that a bank is interested in identifying the individual response of a customer to a certain bid conditional on the customer's characteristics and the loan they have applied for. Such a model could be used for data-driven bid allocations, improved customer segmentation for risk-based pricing, or to support bank clerks in defining the optimal discretionary discount to give to a customer to make a winning bid. To win a customer, two requirements must be met: on the one hand, a bid cannot exceed the overall willingness-to-pay of the customer, that is, even without competition, a customer is not willing to pay any price for the loan. On the other hand,  the loan conditions offered (including the bid) must be more favorable to the customer than those of any of the other competing banks. We note that there might be situations in which a customer might not just pick a bank based on loan conditions but based on personal relations or past experiences. Modeling such a behavior is out of the scope in our study.

In our experiments, we aim to find an unbiased estimate of the individual bid-response function, i.e., the probability of accepting a loan conditional on the bid and customer characteristics:

\begin{equation}
    \mu(b,\mathbf{x}) = \mathbb{E}\left[P(b)|\mathbf{X}=\mathbf{x}\right]
\end{equation}

\noindent To evaluate how a bank could use information about $\mu(b,\mathbf{x})$, we assume that a bank wants to determine the optimal bid $b^{*}_{i}$ per customer that maximizes the expected revenue from the loan pricing and sales process. We expect the revenue $R$ from a loan offer to be proportional to both the height of the bid $b$ and the decision to take out the loan with respect to the bid $Y(b)$:

\begin{equation}
    \label{eq:Revenue}
    R(b) \sim b * Y(b)
\end{equation}

The optimal bid $b_{i}^{*}$ for an individual $i$ is defined as the bid that maximizes the expected revenue for the customer characteristics $\mathbf{x}_i$:

\begin{equation}
    \label{eq:Poptimal}
    \begin{split}
        b^{*}_{i} & = \underset{b}{\arg\max}\text{ } \mathbb{E}\left[ R(b) | \mathbf{X}=\mathbf{x}_{i} \right] \\
        & = \underset{b}{\arg\max}\text{ } b * \mu(b,\mathbf{x}_i)
    \end{split}
\end{equation}

\noindent Our formalization of the loan pricing and sales process makes two simplifications that we want to motivate: first, in reality, a bank is interested in profit over revenue maximization. Yet, to maximize profit, further assessments are needed, such as those of risk. These assessments are beyond the scope of this work. In addition, the bid response might be a component of assessing risk, thereby motivating this simplification. Second, we do not take loan negotiation rounds into account. In several markets, loan pricing and sales processes involve multiple rounds of iterative bidding. We simplify this process and assume that only one bid is made by the bank, as stated before.

\subsection{Selection bias\label{sec:Method_SelecBias}}

\noindent We are interested in learning $\mu(b,\mathbf{x})$ from observational data. This, in itself, is an intricate task, as for every observation, one observes the outcome of a loan pricing and sales process to only the factual bid $b_{f}$ and not to any other. This is known as the ``fundamental problem of causal inference'' \citep[]{Holland.1986}. Furthermore, for a single customer, there is only a small number of observations available, as a single individual takes out only a low number of loans in their lifetime and not necessarily all at the same bank.

\begin{figure}[ht]
    \centering
    \includegraphics[width=0.3\textwidth]{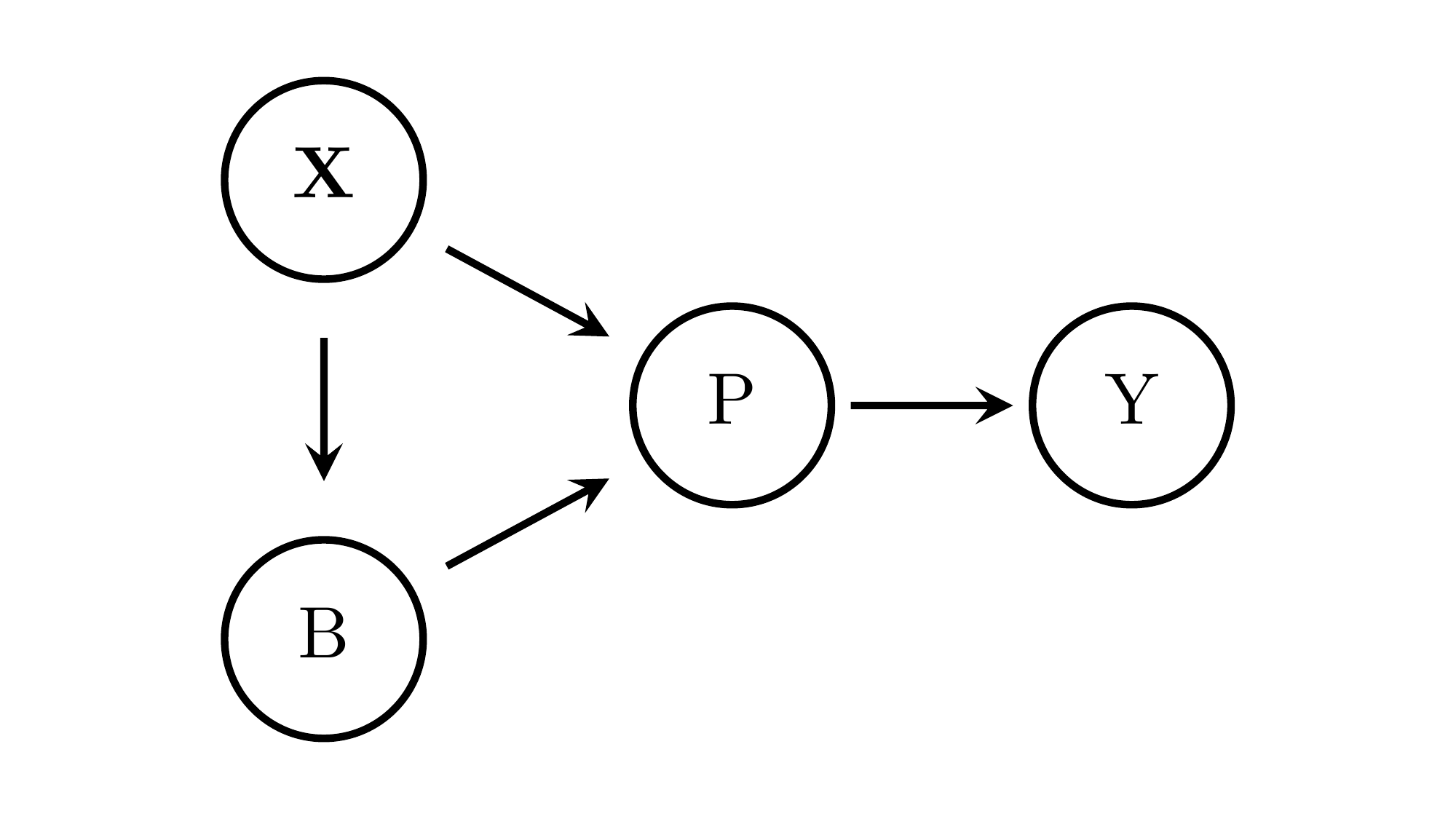}
    \caption{DAG representing the causal dependencies between the variables in the loan pricing and sales process}
    \label{fig:DAG}
\end{figure}

The task is further complicated as the observational data that can be used to learn $\mu$ are expected to be collected under a specific pricing policy, in contrast to data collected using predefined experiments, for example, randomized controlled trials \citep{Rubin.1974, Angrist.2009, Deaton.2018}. This established pricing policy introduces selection bias in the data. Selection bias describes the relationship between the factually assigned bid (or any intervention more generally) and the pretreatment covariates of an observation. Figure~\ref{fig:DAG} shows the dependency of the factual bid and the pretreatment covariates of an observation, as assumed in the remainder of this manuscript. Running randomized control trials in loan pricing is infeasible, for example, due to regulation, compliance, ethical concerns, and costs.

The problem of learning effects of an intervention under the presence of selection bias has been studied for both the case of binary interventions \citep[]{Rosenbaum.1983, Rosenbaum.1984, Yoon.2018} and continuously-valued interventions \citep[]{Imbens.2000, Imai.2004, Hirano.2004, Bica.2020}. Adverse effects of selection bias on traditional approaches to estimating intervention effects have been studied across fields \citep[see, e.g., ][]{Schwab.2020, Vanderschueren.2022}.

We elaborate on three exemplary reasons for bid selection bias in pricing data:
\begin{enumerate}
\item Loan offers were made according to an established pricing policy. Hence, some customer groups are more prone to being assigned a higher or lower bid than others. 
\item Bank clerks might have suffered from implicit stereotypes, either offering higher discretionary discounts to some or even denying loans to certain customer groups \citep{Cozarenco.2018}. 
\item Customers might have suffered from various types of self selection, e.g., driven by socioeconomic status. Customer groups may have a preference for certain loan products over others. Other customer groups might not have applied for a specific loan at all.
\end{enumerate}

\subsection{Assumptions\label{sec:Method_Assumptions}}

\noindent In the literature, we find three assumptions that  typically underly the estimation of intervention effects from observational data \citep[]{Imbens.2000, Lechner.2001}. We will discuss these assumptions and their applicability in pricing:

\begin{assumption}
\textbf{Overlap:} $\forall b \in \mathbb{R}: \forall \mathbf{x} \in \mathbb{R}^d \text{ with } \mathbb{P}(\mathbf{x}) > 0 : 0 < \mathbb{P}(b|\mathbf{x}) < 1$
\end{assumption}
\noindent Overlap requires that for every possible customer characteristic $\mathbf{x}$, every possible bid $b$ has a nonzero possibility of being offered. In reality, an established pricing policy might assign specific bids with little to no deviation to customer groups, resulting in a violation of the overlap assumption. Overlap might be ensured, however, due to bank clerks assigning discretionary discounts of sufficient size to their customers. The dispersion of bids ensures that the true bid response can be estimated for all possible bid levels for a certain customer.

\begin{assumption}
\textbf{Consistency:} $\forall \mathbf{x} \in \mathbb{R}^{d}: Y_{f} = (Y | \mathbf{X}=\mathbf{x})$
\end{assumption}
\noindent Consistency means that the observed outcome  is the true potential outcome for the applied treatment. As \citet{Vanderschueren.2022} note, this appears to be a straightforward assumption. To ensure consistency in pricing, we must carefully observe any changes in economic sentiment and correct or retrain models if, e.g., prime rates change. For example, if monetary policy tightens, the response of a customer to a certain bid $b$ will likely change as well. For our experiments, we  assume that the economic sentiment is stable and that consistency can be assumed.

\begin{assumption}
\textbf{Unconfoundedness:} $\left(P(b)|b \in \mathbb{R}\right) \perp\!\!\!\!\perp B_f|\mathbf{X}$
\end{assumption}

\noindent Unconfoundedness requires that all information that determined the factually assigned bid is known and available for modeling. In pricing, this assumption is easily violated, e.g., due to unrecorded interactions between bank clerks and customers \citep{Phillips.2012a}, legal bounds on the use of personal data, and missing information about competitor offers, commonly referred to as ``competitive uncertainty''.\\

\noindent For the purpose of our work, we do not further evaluate the impacts of violating these assumptions on estimating bid responses. Instead, we assume the assumptions are met, to isolate the effect of selection bias.

\subsection{Predicting bid-response curves\label{sec:Method_Predict}}

\noindent We  evaluate a total of seven methods to model individual bid responses, covering parametric models, nonparametric models, and causal and noncausal methods, as depicted in Table~\ref{tbl:Methods}. Each method is discussed further below:

\begin{table}[ht]
    \centering
    \caption{Models for bid-response prediction}
    \label{tbl:Methods}
    \footnotesize
    \centering
    \begin{tabular}[c]{lcc}
        \hline 
        \hline \\[-2.0ex] 
        Method & Parametric & Causal \\
        \hline \\[-1.0ex]
        Na\"ive pricing & n.a. & \tikzxmark \\
        Logistic regression & \tikzcmark & \tikzxmark \\
        Random Forest & \tikzxmark & \tikzxmark\\
        MLP & \tikzxmark & \tikzxmark\\
        HIE & \tikzcmark & \tikzcmark\\
        VCNets & \tikzxmark & \tikzcmark\\
        DRNets & \tikzxmark & \tikzcmark\\[-0.1ex]
        \hline\\[-1.5ex]
    \end{tabular}\\
    n.a.: not applicable
\end{table}

We consider four noncausal methods for bid-response learning to test their robustness to selection bias and appropriateness to modeling bid-response functions: first, na\"ive pricing is a strategy in which we do not estimate any dose response but assume that the best bid $b^{*}$ per customer is the factually assigned bid under the established pricing policy. Second, logistic regression is a classification method  that is widely adopted across scientific domains and industry applications, such as fraud detection, credit risk modeling, and churn prediction. It is also the de facto standard in estimating bid-response models \citep{vanRyzin.2012} due to its well-suited hypothesis space. Logistic regression is fully parametric and easy to interpret. Third, we consider a random forest classifier \citep{Breiman.2001}. The random forest classifier combines multiple decision trees into an ensemble  to improve the accuracy and reduce the variance of the final model. Random forests can handle a variety of input variables and are robust to overfitting. Random forests, and tree-based methods more generally, have found wide adoption in industry applications of machine learning, motivating our choice to adopt them in our experiments. Fourth, we fit an artificial neural network, i.e., a multilayer perceptron (MLP). MLPs are universal function approximators \citep{Hornik.1989} and are powerful methods for a multitude of prediction tasks, making them an important benchmark for our experiments. 

\begin{figure}[h]
    \centering
    \begin{subfigure}{0.45\textwidth}
        \centering
        \includegraphics[width=1.0\textwidth]{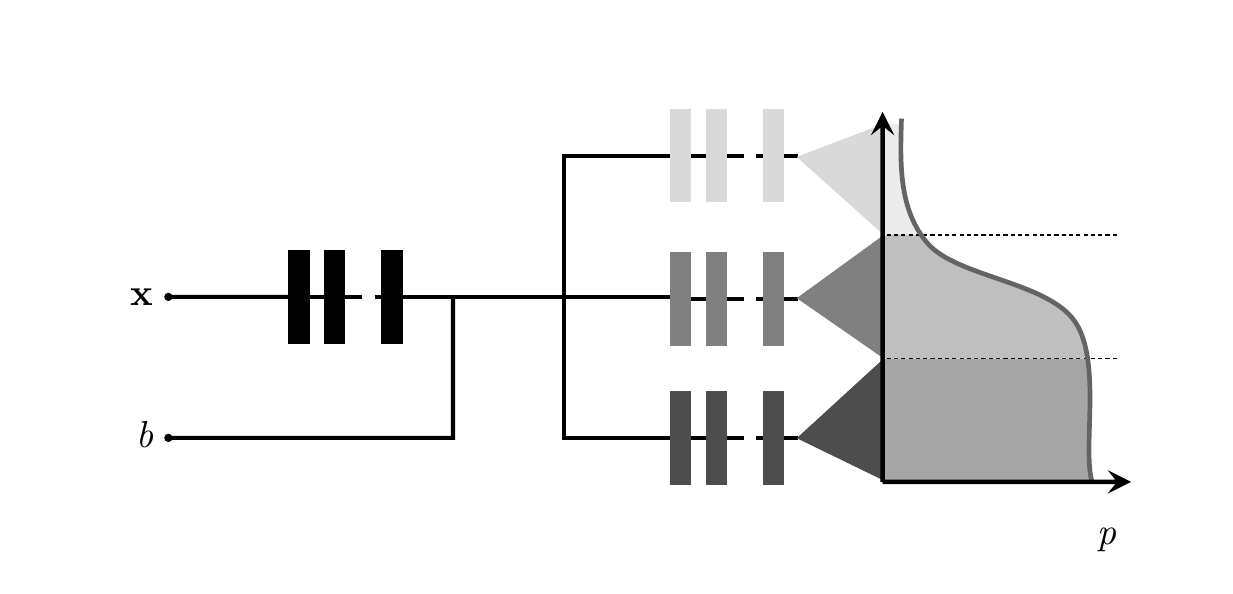}
        \caption{DRNets}
        \label{fig:DRNets}
    \end{subfigure}
    \hfill
    \begin{subfigure}{0.45\textwidth}
        \centering
        \includegraphics[width=1.0\textwidth]{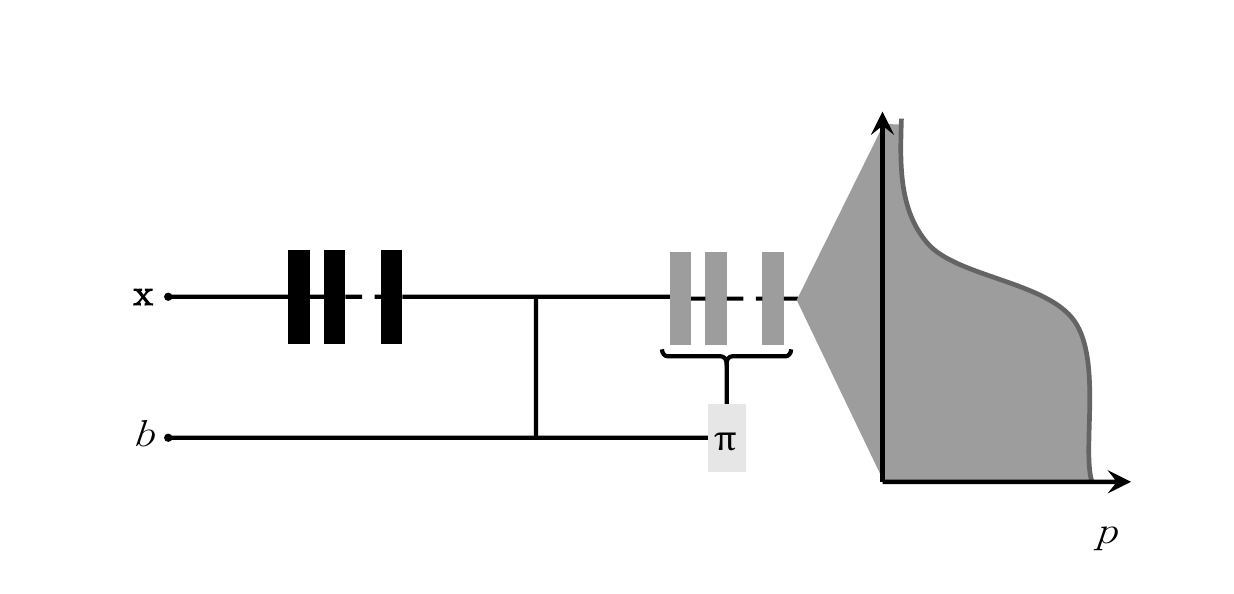}
        \caption{VCNets}
        \label{fig:}
    \end{subfigure}
    \caption{Network architectures}
    \label{fig:Architectures}
\end{figure}

Additionally, we implement three causal methods that are meant to handle selection bias by design, expecting that these methods will show improved robustness to increasing levels of bid selection bias in pricing data. First, we implement the Hirano Imbens estimator \citep[``HIE'', ][]{Hirano.2004}, which builds on the idea of the generalized propensity score, as proposed by \citet{Imbens.2000}. The HIE is an imputation-type estimator. Initially, a model of the treatment assignment is built. Based on that model, the propensity for every observation and their factual bid is calculated. Finally, the bid response is modeled as a function of the bid level and the propensity score. We follow \citet{Schwab.2020} and \citet{Bica.2020} in considering a linear relationship between bid and pretreatment covariates with normally distributed error terms, as well as a logistic regression on the second-order polynomial of bid level and propensity score as the final model of the bid response. Second, we implement DRNets \citep{Schwab.2020}. DRNet extends TARNet \citep{Shalit.2017} for estimating effects of binary interventions to the continuous case, by training a neural network with $\rho$ individual prediction heads for observations with different levels of a continuous intervention. DRNets learn a shared representation of the training observations, which  implicitly have a regularizing effect on overfitting the training data. Third, we implement VCNets. \citet{Nie.2021} propose VCNets as an extension to DRNets, by proposing a varying coefficient neural network, where coefficients of the neural network are a function of the continuously valued intervention, enforcing a continuous dose response function, over the tendency of DRNets to estimate discontinuous functions. We use the architecture proposed in \citet{Nie.2021}. We illustrate the architectures of DRNets and VCNets in Figure~\ref{fig:Architectures}. Finally, we would like to motivate our choice of not including SCIGAN \citep[]{Bica.2020},  another powerful causal machine learning method for the estimation of continuously valued interventions, in the experiments. Even though SCIGAN has shown high performance in previous studies \citep[]{Bica.2020, Vanderschueren.2022}, the set-up of using generative models for the generation of counterfactuals to learn unbiased responses is not suited for the case of pricing, as the training data are not continuous. In our own experiments, SCIGAN performed worse than both DRNets and all other methods considered.

\section{Experimental Evaluation\label{sec:Experiments}}

\subsection{Data\label{sec:Experiments_Data}}

\noindent We use a real-life dataset on mortgage loan offers from a bank in Belgium. The data cover over 12,000 loan offers. Each offer relates to precisely one customer. In collaboration with the bank, we selected and processed an anonymized subset of 13 variables for training. The selection of variables was based on their availability for all customers to ensure no missing data, an expert assessment of their relevance for estimating price sensitivity. Categorical variables have been dummy encoded, and all variables have been standardized. These variables include information on terms and conditions of the loan (5 variables), financial background of the customer (3 variables), the socioeconomic background of the customer (4 variables), and the existing relationship of the customer with the bank (1 variable). The precise definition of the variables is confidential information that cannot be shared.

As highlighted in Section~\ref{sec:Method_Problem}, evaluating the correctness of predicted potential outcomes is challenging. First, we never observe the true bid-response curve but only the outcome to one bid per customer due to the fundamental problem of causal inference. Second, estimating the overall effectiveness of one pricing policy over another over a group of customers is complicated due to the observational data. Methods designed to estimate such an aggregate effect, such as the Qini-coefficient \citep{Radcliffe.2007}, require randomized controlled trial (RCT) data \citep[]{Angrist.2009} and are negatively affected by selection bias. Due to compliance, regulation, and the risks of losses  and reputation, such trials cannot be implemented reasonably in many real-life settings, including banking operations.

To overcome this issue, we make use of semisynthetic data, a common approach in evaluating potential outcome estimation and to avoid the fundamental problem of causal inference \citep[][]{Yoon.2018, Berrevoets.2020, Schwab.2020, Qian.2023}. In such a setup, typically, the explanatory variables in the data are kept as-is, whereas both treatment variables $B$ and outcomes $P$ and $Y$ are generated from a well-defined and known ground truth to control  factors such as the strength of the selection bias and the functional form of the potential outcome in relation to $\mathbf{X}$. We deem the use of semisynthetic data in pricing as particularly interesting to assess the impacts of selection bias on modeling bid responses.

In the following two subsections, we discuss how we model the underlying ground-truth bid response (Section~\ref{sec:Experiments_Data_GroundTruth}) and assignment of factual bids (Section~\ref{sec:Experiments_Data_SelectBias}). For our experiments, we use 70\% of the total data for training, 10\% for validation and parameter tuning and 20\% for testing. 

\subsubsection{Ground-truth bid response\label{sec:Experiments_Data_GroundTruth}}

\noindent Based on \citet{vanRyzin.2012} and \citet{Phillips.2021}, we define four requirements with regards to a synthetic ground-truth bid-response function. In our experiments, we standardize bids to be in the domain $[0,1]$, with $b=0$ the lowest observed bid and $b=1$ the highest observed bid:

\begin{itemize}
    \item At $b=0$, every customer has a unique probability of acceptance, $p(0,\mathbf{x}) \leq 1$.\\
    The probability is $p \leq 1$ but not $p = 1$, as even for the lowest observed bid, a customer might consider the bid to be too high according to their individual preferences or reject a bid based on their customer experience during the loan pricing and sales process.
    \item When $b$ increases, the probability of accepting a loan converges but not necessarily towards zero, i.e., $p(\infty,\mathbf{x}) \geq 0$.\\
    The probability does not necessarily converge to zero, as even for the highest observed bid, a customer might take out a loan.
    \item The price sensitivity, that is, $\frac{\text{d}p(b,\mathbf{x})}{\text{d}b}$, is heterogeneous across customers.\\
    Ceteris paribus, two customers might react differently in terms of their bid response to an increase of the bid.
    \item The probability of loan acceptance is monotonically decreasing for an increasing value of the bid, $b$.
\end{itemize}

\begin{figure}[t]
    \centering
    \begin{subfigure}{0.45\textwidth}
        \centering
        \includegraphics[width=0.75\textwidth]{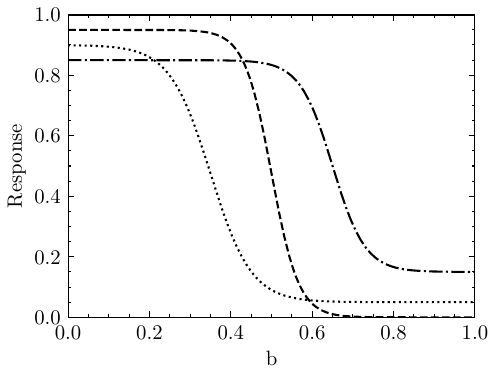}
        \caption{Richards curves}
        \label{fig:price_responses_rc}
    \end{subfigure}
    \hfill
    \begin{subfigure}{0.45\textwidth}
        \centering
        \includegraphics[width=0.75\textwidth]{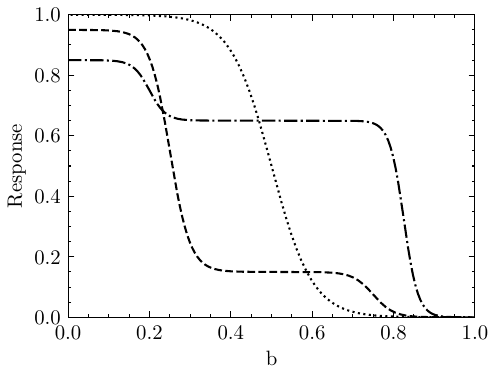}
        \caption{Stacked sigmoid curves}
        \label{fig:price_responses_ss}
    \end{subfigure}
    
    \caption{Illustrative visualization of bid-response curves}
    \label{fig:price_responses}
\end{figure}

\noindent Furthermore, we consider two different ground-truth bid-response curves: a Richards curve and a stacked sigmoid curve (Figure~\ref{fig:price_responses} provides a comparative visualization of both curves).\\

\noindent \textbf{Richards curve:} The generalized logistic function or Richards curve \citep[]{Richards.1959} is defined as:
\begin{equation}
    \label{eq:GroundTruth}
    p(b,\mathbf{x})_{RC} = (1 - \alpha) - \frac{\beta-\alpha}{1+e^{-\gamma * (\delta - b)}} + \epsilon
\end{equation}
with $1 - \alpha$, the left asymptote (or maximum level of acceptance at the lowest observed bid $b_{min}$), $1 - \beta$, the right asymptote (or minimum level of acceptance at the highest observed price $b_{max}$), $\gamma$ the steepness of the curve and $\delta$ the position of the turning point of the curve. The logistic  function has seen various applications across domains and is often used as a modeling tool for classification tasks \citep[]{Cramer.2002}.

To condition $p(b,\mathbf{x})_{RC}$ on individual customer characteristics, we parameterize the function by means of linear combinations of the input features in $\mathbf{x}$ via:

\begin{equation}
\begin{aligned}
    \alpha  & = (0.2 \mathbf{w}_{1}^{\intercal}\mathbf{x}) \\
    \beta & = (0.8 + 0.2 (\mathbf{w}_{2}^{\intercal}\mathbf{x})) \\
    \gamma & = (0.5 + 5 (\mathbf{w}_{3}^{\intercal}\mathbf{x})) \\
    \delta & = (\mathbf{w}_{4}^{\intercal}\mathbf{x}) \\
    \epsilon & \sim \mathcal{N}(0, 0.1)
\end{aligned}
\end{equation} 
For all $i \in \{1,2,3,4\}$, $\mathbf{w}_{i} \sim \mathcal{U}((0,1)^{d \times 1})$, as proposed in \citet{Bica.2020}.\\

\noindent \textbf{Stacked sigmoid:} To test the performance of models in the case of highly nonlinear bid-response curves, we evaluate them considering the bid response as a mixture of two sigmoid curves. We define:

\begin{equation}
    \label{eq:GroundTruth_SS}
    \begin{aligned}
        p(b,\mathbf{x})_{SS} = \alpha + \beta * \sigma(P/\gamma) + (1 - \alpha - \beta) * \sigma((b-\delta)/(1-\delta)) + \epsilon
    \end{aligned}
\end{equation}
$\sigma(x)$ is defined as a sigmoid curve:
\begin{equation}
    \label{eq:std_sigmoid}
        \sigma(x) = 1 - \frac{1}{1+e^{-20*(0.5 - x)}}
\end{equation}
As before, we condition $p(b,\mathbf{x})_{SS}$ on individual customer characteristics via:
\begin{equation}
\begin{aligned}
    \alpha  & = (0.2 \mathbf{w}_{1}^{\intercal}\mathbf{x}) \\
    \beta & = (0.8 \mathbf{w}_{2}^{\intercal}\mathbf{x}) \\
    \gamma & = \mathbf{w}_{3}^{\intercal}\mathbf{x} \\
    \delta & = \mathbf{w}_{4}^{\intercal}\mathbf{x} \\
    \epsilon & \sim \mathcal{N}(0, 0.1)
\end{aligned}
\end{equation} 
For all $i \in \{1,2,3,4\}$, $\mathbf{w}_{i} \sim \mathcal{U}((0,1)^{d \times 1})$.

\subsubsection{Bid assignment\label{sec:Experiments_Data_SelectBias}}

\begin{figure}[t]
    \centering
    \begin{subfigure}{0.3\textwidth}
        \centering
        \includegraphics[width=\textwidth]{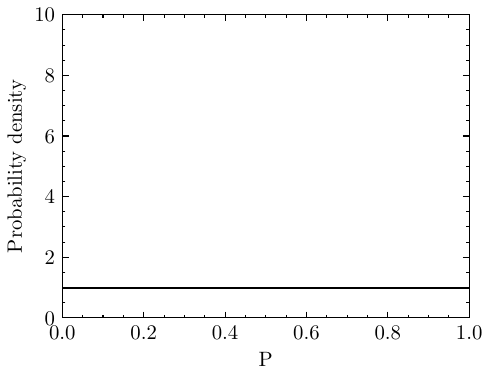}
        \caption{$\theta = 0$}
        \label{fig:selectionbias_0}
    \end{subfigure}
    \hfill
    \begin{subfigure}{0.3\textwidth}
        \centering
        \includegraphics[width=\textwidth]{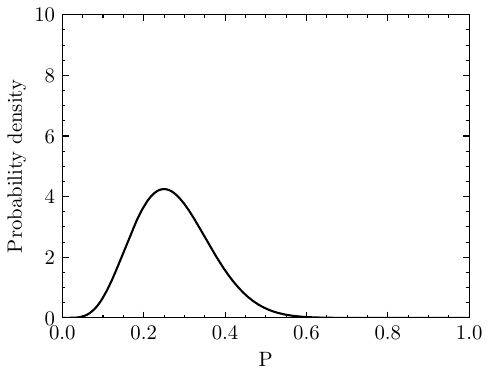}
        \caption{$\theta = 5$}
        \label{fig:selectionbias_5}
    \end{subfigure}
    \hfill
    \begin{subfigure}{0.3\textwidth}
        \centering
        \includegraphics[width=\textwidth]{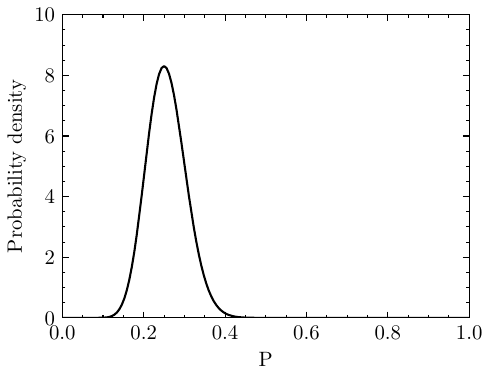}
        \caption{$\theta = 20$}
        \label{fig:selectionbias_20}
    \end{subfigure}
    
    \caption{Visualization of different levels of selection bias $\theta$ for mode $\mu = 0.25$}
    \label{fig:selectionbias}
\end{figure}

\noindent Next to the simulation of the bid response, we control the levels of selection bias by sampling the factual bid $b_{f}$ from a beta distribution. The approach is motivated and detailed in \citet{Bica.2020}. We model varying levels of selection bias by assigning to every observation $\mathbf{x}$ a factual bid:
\begin{equation}
    b_{f} \sim Beta \left( \theta + 1, \frac{\theta}{\phi + (1 - \theta)} \right)
\end{equation}
where $\theta \geq 1$ controls the level of selection bias. $\theta = 0$ results in no selection bias. $\mu$ defines the modal rate, i.e., the most likely assigned price, and is defined as 
\begin{equation}
    \phi = \mathbf{w}_{5}^{\intercal}\mathbf{X}
\end{equation} 
with $\mathbf{w}_{5} \sim \mathcal{U}((0,1)^{d \times 1})$. Hence, the distribution of the assigned bids in the dataset depends on $\mathbf{X}$, thereby introducing selection bias. Different levels of bias are illustrated in Figure~\ref{fig:selectionbias}.

Finally, in addition to directly controlling the level of selection bias, we use the observed bid levels from the original dataset to assess the impact of levels of bid selection bias observed in real-world scenarios.

\subsection{Evaluation metrics\label{sec:Experiments_EvalMetrics}}

\noindent We evaluate our method using three performance metrics:\\

\noindent \textbf{Mean Integrated Squared Error (MISE)}: The MISE evaluates the potential of a method to predict the true bid response per customer along all observed bid levels:

\begin{equation}
    \label{eq:MISE}
    \text{MISE} = 
    \frac{1}{n} \sum_{i = 1}^{n} \int_{b_{min}}^{b_{max}} \left(\mu(b,\mathbf{x}_i) - \hat{\mu}(b,\mathbf{x}_{i}) \right)^{2} db
\end{equation}
with $n$ the number of test observations, $b_{min}$ the lowest bid, and $b_{max}$ the highest bid observed in the training data\footnote{Due to the overlap assumption (cf. Section~\ref{sec:Method_Assumptions}), we only evaluate bids  between $b_{min}$ and $b_{max}$}. The MISE was initially proposed for individual continuous treatment effect estimation by \citet[]{Schwab.2020}\footnote{In addition to to evaluating the bid response, we evaluate the MISE over the expected revenue in accordance with Equation~\ref{eq:Poptimal}, further referred to as the \textit{MISE R}. The results can be found in the Appendix.}.\\

\noindent \textbf{Policy error (PE)}: The PE evaluates how well a method is able to identify the optimal bid $b^{*}$ and was proposed by \citet{Schwab.2020}:

\begin{equation}
    \label{eq:PE}
    \text{PE} = \frac{1}{n} \sum_{i = 1}^{n} \left( b_{i}^{*} - \Hat{b_{i}^{*}} \right)^2
\end{equation}
with $n$ the number of test observations, $b_{i}^{*}$ the actually optimal bid of observation $i$, and $\Hat{b_{i}^{*}}$ the estimated optimal bid.\\

\noindent \textbf{Brier score (BS):} Finally, we evaluate each model in terms of their Brier score \citep[]{Brier.1950} to assess the potential to estimate the factual outcome of a loan pricing and sales process. The Brier score informs about a model's capacity to estimate the factual response:

\begin{equation}
    \label{eq:Brierscore}
    \text{BS} = 
    \frac{1}{n} \sum_{i = 1}^{n} \left( y_{i,f} - \hat{\mu}(b_{i,f},\mathbf{x}_{i}) \right)^2
\end{equation}
with $y_{i,f}$ the factual outcome of observation $i$ and $\hat{\mu}(b_{i,f},\mathbf{x}_{i}$ the estimated probability of bid acceptance of observation $i$ for their factually assigned bid.

\subsection{Implementation\label{sec:Experiments_Implementation}}

\noindent We implement our experiments in Python 3.9. Neural network-based methods are implemented in PyTorch \citep{Paszke.2017}. Random forest models and logistic regression are implemented using the SciKit-Learn library \citep[]{Scikit.2011}. HIE is implemented using the statsmodels library \citep[]{Seabold.2010}. For VCNets, we use the implementation provided in the original work of \citet{Nie.2021}. For a list of hyperparameters used to train each method, we refer to Section~\ref{sec:Append_Hyperparam} in the Appendix. All code used for this project is available online at:
\begin{equation*}
    \texttt{https://github.com/christopher-br/Causal\_Pricing}.
\end{equation*}

\subsection{Experimental results\label{sec:Experiments_Results}}

\noindent We apply our data generating process to the data presented in Section~\ref{sec:Experiments_Data}. We run the process a total of 10 times for each of seven bias strengths increasing from no bias ($\theta = 0$) to heavy bias ($\theta = 20)$, as well as the factual bids. The stated performance metrics are averaged over each of the 10 randomly initialized runs.

The metrics introduced in Section~\ref{sec:Experiments_EvalMetrics} enable evaluation of each model in terms of its capacity to predict the bid response along all observed bids (via the MISE metric) and its appropriateness for operational decision-making to set an optimal bid $b^{*}$ (via the PE metric). The Brier score, moreover, evaluates the ability to estimate the factual outcome of an observation.

\begin{table}[b]
    \centering
    \caption{Mean Integrated Squared Error (MISE) on Richards curve}
    \footnotesize
    \begin{tabular}{lccccccc:c}
        \multicolumn{9}{l}{\textbf{Ground truth:} Richards curve}\\
            \hline\\[-2.3ex]
	    \hline &&&&&&&&\\[-2.0ex] 
	    & \multicolumn{7}{c:}{\textit{Bias strength}} & \multirow{2}{*}{True bids}\\
	    \cdashline{2-8}&&&&&&&&\\[-2.0ex]
        Model & 0.0 & 2.5 & 5.0 & 7.5 & 10.0 & 15.0 & 20.0 & \\
        \hline &&&&&&&&\\[-1.0ex]
        Na\"ive pricing & 	n.a. & 	n.a. & 	n.a. & 	n.a. & 	n.a. & 	n.a. & 	n.a. & 	n.a.	\\
        Logistic Regression & 	0.115 & 	0.095 & 	0.096 & 	0.097 & 	\textit{0.100} & 	\textit{0.104} & 	\textit{0.112} & 	0.141	\\
        Random Forest & 	0.083 & 	0.105 & 	0.122 & 	0.135 & 	0.143 & 	0.152 & 	0.170 & 	0.164	\\
        MLP & 	\textbf{0.056} & 	\textbf{0.062} & 	\textbf{0.058} & 	\textbf{0.062} & 	\textbf{0.066} & 	\textbf{0.060} & 	\textbf{0.064} & 	\textbf{0.068}	\\
        \hdashline&&&&&&&&\\[-2.0ex]						
        HIE & 	0.118 & 	0.120 & 	0.122 & 	0.124 & 	0.124 & 	0.125 & 	0.126 & 	0.126	\\
        DRNets & 	\textit{0.076} & 	\textit{0.082} & 	0.117 & 	0.117 & 	0.111 & 	0.118 & 	0.140 & 	\textit{0.111}	\\
        VCNets & 	0.088 & 	\textit{0.082} & 	\textit{0.084} & 	\textit{0.092} & 	\textit{0.100} & 	\textit{0.104} & 	0.114 & 	0.112	\\
        \hline \\[-1.0ex]
    \end{tabular}\\
    \footnotesize n.a.: not applicable
    \label{tbl:Results_MISEs}
\end{table}

Note that the stated performances depend on both our data generation process (cf. Section~\ref{sec:Experiments_Data}) and the operational objective we have set (cf. Section~\ref{sec:Method_Problem}).\\

\noindent \textbf{Dose response prediction (MISE, Tables~\ref{tbl:Results_MISEs}~and~\ref{tbl:Results_MISEs_SS}):} If the true dose-response function assumes the shape of a Richards curve (cf. Table~\ref{tbl:Results_MISEs}), the standard neural network (MLP) achieves the best performance. Second best are standard logistic regression and VCNets. Contrarily to what we had initially assumed, logistic regression and the MLP are seemingly least affected by the increase in selection bias, whereas, e.g., DRNets (a causal method) sees an increase of almost 100\% in MISE from no bias to the strongest bias.

The strong performance of logistic regression is likely explained by its hypothesis space being closely related to the ground-truth bid response in the simulation. As the hypothesis spaces are comparable, standard logistic regression is likely able to find a well-performing model by averaging over the observed data. Likewise, DRNet is not able to adequately fit the dose responses. We relate this to the criticism of \citet{Nie.2021} on DRNets being prone to fitting discontinuous functions due to estimating separate head networks for different bid levels. This can explain why VCNets improve upon DRNets in the case of the Richard's curve, as their estimated bid-response curve is continuous. HIE does not improve upon the nonparametric causal methods, which might be due to the functional form chosen for modeling the treatment assignment or bid response.

\begin{table}[ht]
    \centering
    \caption{Mean Integrated Squared Error (MISE) on stacked sigmoid}
    \footnotesize
    \begin{tabular}{lccccccc:c}
        \multicolumn{9}{l}{\textbf{Ground truth:} Stacked sigmoid}\\
            \hline\\[-2.3ex]
	    \hline &&&&&&&&\\[-2.0ex] 
	    & \multicolumn{7}{c:}{\textit{Bias strength}} & \multirow{2}{*}{True bids}\\
	    \cdashline{2-8}&&&&&&&&\\[-2.0ex]
        Model & 0.0 & 2.5 & 5.0 & 7.5 & 10.0 & 15.0 & 20.0 & \\
        \hline &&&&&&&&\\[-1.0ex]
        Na\"ive pricing & 	n.a. & 	n.a. & 	n.a. & 	n.a. & 	n.a. & 	n.a. & 	n.a. & 	n.a.	\\
        Logistic Regression & 	0.110 & 	0.125 & 	0.134 & 	0.143 & 	0.147 & 	0.154 & 	0.155 & 	\textit{0.119}	\\
        Random Forest & 	0.088 & 	0.128 & 	0.151 & 	0.170 & 	0.174 & 	0.186 & 	0.196 & 	0.196	\\
        MLP & 	\textbf{0.059} & 	\textbf{0.073} & 	\textit{0.100} & 	\textit{0.110} & 	\textit{0.119} & 	\textit{0.135} & 	\textit{0.147} & 	0.132	\\
        \hdashline&&&&&&&&\\[-2.0ex]							
        HIE & 	0.087 & 	0.101 & 	0.109 & 	0.122 & 	0.121 & 	0.149 & 	0.157 & 	0.176	\\
        DRNets & 	\textit{0.075} & 	\textit{0.080} & 	\textbf{0.098} & 	\textbf{0.093} & 	\textbf{0.115} & 	\textbf{0.128} & 	\textbf{0.109} & 	\textbf{0.103}	\\
        VCNets & 	0.099 & 	0.111 & 	0.136 & 	0.150 & 	0.156 & 	0.172 & 	0.178 & 	0.120	\\
        \hline \\[-1.0ex]
    \end{tabular}\\
    \footnotesize n.a.: not applicable
    \label{tbl:Results_MISEs_SS}
\end{table}

If we assume higher degrees of nonlinearity in the dose response, i.e., in the case of a stacked sigmoid ground truth (cf. Table~\ref{tbl:Results_MISEs_SS}), we observe an increasingly adverse effect of selection bias on estimating the bid response. The logistic regression model is not able to model the functional shape of the bid response with an increasing MISE for higher levels of selection bias. The random forest classifier is again most affected by selection bias, performing worst of the methods we have applied. As with the Richard's curve, the MLP achieves superior performance when no selection bias is present. This time, when bias increases, the MLP is significantly affected by the selection bias in the data. In this setting, DRNets exceed the performance of other methods, generating second-best results for low levels of bias and achieving best results for bias levels larger than $\theta = 2.5$, as well as for the bid levels observed in the original dataset. DRNets even outperform VCNets, indicating their applicability to causal effect estimation when the training data are binary. VCNets appear to be not well suited in these kinds of settings, as soon as there is a larger degree of nonlinear relations in the data.\\

\noindent \textbf{Optimal bid prediction (PE, Tables~\ref{tbl:Results_PEs}~and~\ref{tbl:Results_PEs_SS}):} In terms of policy error, we observe similar results as with the MISE. If the true dose response is a Richards curve (cf. Tabel~\ref{tbl:Results_PEs}), the best-performing models in terms of PE are the MLP and logistic regression. The added flexibility of nonparametric methods, or the use of causal methods, is not sufficient to outperform these two baseline approaches. Again, the method seemingly most affected by selection bias is the random forest classifier.

With increased nonlinearity, the causal methods again score better and best. For moderate levels of bias, VCNets outperform DRNets, likely due to their continuous estimate of the bid response over the discountinuous function of DRNets. If the bias increases further, this benefit is not sufficient to overcome the performance of DRNets, seemingly a more robust estimator.\\

\noindent \textbf{Factual outcome estimation (BS, Tables~\ref{tbl:Results_BSs}~and~\ref{tbl:Results_BSs_SS}):} Evaluating each of the methods in terms of the Brier score illustrates the need to look at pricing from a causal perspective. If models are solely tested on their predictive performance within the established pricing mechanism, the results are robust across all levels of selection bias. Most noteworthy, the random forest classifier achieves second-best performance under the highest level of selection bias (cf. Table~\ref{tbl:Results_BSs}), even though it achieves the lowest performance in terms of MISE and PE. This finding stresses that the Brier score cannot be used to evaluate how well a pricing model can generalize across bid levels, i.e., its generalization power. Hence, testing and selecting models based on the Brier score might even introduce vicious cycles \citep[see, e.g., ][]{Mansoury.2020, Adam.2020}, an important threat that must be further investigated in the case of banking and lending operations.

\begin{table}[ht]
    \centering
    \caption{Brier score (BS) on Richards curve}
    \footnotesize
    \begin{tabular}{lccccccc:c}
        \multicolumn{9}{l}{\textbf{Ground truth:} Richards curve}\\
            \hline\\[-2.3ex]
	    \hline &&&&&&&&\\[-2.0ex] 
	    & \multicolumn{7}{c:}{\textit{Bias strength}} & \multirow{2}{*}{True bids}\\
	    \cdashline{2-8}&&&&&&&&\\[-2.0ex]
        Model & 0.0 & 2.5 & 5.0 & 7.5 & 10.0 & 15.0 & 20.0 & \\
        \hline &&&&&&&&\\[-1.0ex]
        Na\"ive pricing & 	n.a. & 	n.a. & 	n.a. & 	n.a. & 	n.a. & 	n.a. & 	n.a. & 	n.a.	\\
        Logistic Regression & 	0.355 & 	0.366 & 	0.371 & 	0.377 & 	0.372 & 	0.372 & 	0.372 & 	0.342	\\
        Random Forest & 	\textit{0.345} & 	0.362 & 	0.367 & 	0.375 & 	0.370 & 	0.368 & 	\textit{0.368} & 	\textit{0.337}	\\
        MLP & 	\textbf{0.342} & 	\textbf{0.356} & 	\textbf{0.362} & 	\textbf{0.369} & 	\textbf{0.362} & 	\textbf{0.362} & 	\textbf{0.362} & 	\textbf{0.334}	\\
        \hdashline&&&&&&&&\\[-2.0ex]									
        HIE & 	0.356 & 	0.381 & 	0.389 & 	0.396 & 	0.394 & 	0.392 & 	0.389 & 	0.345	\\
        DRNets & 	\textit{0.345} & 	\textit{0.359} & 	\textit{0.366} & 	\textit{0.373} & 	\textit{0.368} & 	\textit{0.367} & 	0.369 & 	\textit{0.337}	\\
        VCNets & 	0.347 & 	0.362 & 	0.367 & 	0.374 & 	0.369 & 	0.368 & 	0.369 & 	0.341	\\
        \hline \\[-1.0ex]
    \end{tabular}\\
    \footnotesize n.a.: not applicable
    \label{tbl:Results_BSs}
\end{table}

\section{Conclusion\label{sec:Conclusion}}

\noindent In this paper, we explore the use of (causal) machine learning to estimate models of individual bid-response curves from observational loan pricing data. We investigate the adverse effects of bid selection bias. In a series of experiments, we evaluate the impact of varying levels of bid selection bias using a semisynthetic dataset on mortgage loan applications and test several data-driven approaches in learning individual bid-response models. Thereby, we frame pricing as problem of causal inference and, more specifically, of individual treatment effect estimation.

The presented results show that ensuring robustness to bid selection bias is crucial in developing a data-driven pricing solution, which potentially is an overlooked issue and threat both in literature and industry. Failing to adjust for selection bias may significantly impact bid-response models that are obtained from observational data and, as such, introduce systematic errors. This, in turn, might induce vicious feedback cycles in which the effectiveness of a pricing policy is both overestimated and suboptimal. Tree-based models, such as random forests, seem especially unable to accurately predict counterfactual outcomes, even when their performance on estimating the outcomes of an existing pricing policy might be convincing. The wide adoption of tree-based methods in the industry makes this a substantial threat that calls for further research. Finally, we show that approaches from causal machine learning prove potent in overcoming issues that result from selection bias, at least when the underlying assumptions discussed in this paper are satisfied. 

We draw two key conclusions from our seminal work:

\begin{enumerate}
    \item The estimation of bid responses from observational data varies from previously discussed settings of estimating dose responses \citep[]{Schwab.2020, Bica.2020, Vanderschueren.2022}. In estimating a bid response, practitioners must infer a continuous function from binary training data. This significantly complicates the task in comparison with previous applications with continuous training data. Methodologies such as VCNet have not been able to show the same level of robustness to selection bias as in preceding studies, e.g., as in \citet{Nie.2021}.
    \item Traditional machine learning methods suffer from increasing levels of bid selection bias, especially when the degree of nonlinearity in the underlying ground-truth bid response increases. Causal machine learning methods can help to overcome this issue but cannot not fully solve it yet.
\end{enumerate}

Having adopted a semisynthetic set of experiments, we want to highlight the assumptions and limitations of our work: in our data simulation (cf. Section~\ref{sec:Experiments_Data}), we purposefully satisfy both overlap and unconfoundedness, two assumptions that are deemed critical for the application of causal inference. Both assumptions might be violated in real-life data. We currently assume that the decision to accept a certain bid is a function of only the observed characteristics $\mathbf{X}$ and the bid $B$. This does not hold in reality, where hidden confounding is likely to occur, as explained in Section~\ref{sec:Method_Problem} and prominently discussed in previous studies on endogeneity in pricing (cf. Section~\ref{sec:LitRev}). We see two interlinking components: first, customers compare bid offers obtained from different banks, which are typically not observed. Second, some available data are not permitted to be used for business decision making, for example, under regulation 2016/679 of the European Parliament\footnote{Regulation (EU) 2016/679 of the European Parliament and of the Council of 27 April 2016 on the protection of natural persons with regard to the processing of personal data and on the free movement of such data, and repealing Directive 95/46/EC (General Data Protection Regulation) [2016] OJ L 119/1.} (GDPR). Finally, our experiments do not consider any temporal component in the training data. If economic sentiment changes and average credit rates decrease, the bid response, $\mu(b, \mathbf{x})$, ceteris paribus, will certainly shift, but possibly change in form as well.

We conclude with an outlook on future work. First, we believe that developing and testing causal machine learning methods for business processes such as pricing is crucial but has not yet been tackled in the established literature. Specifically, we call for the development of methods that learn an individual probability estimate as an effect on a continuously valued intervention from binary training data. This research might advance the application of causal machine learning in many other fields, such as risk assessment and policy evaluation. Second, further research should evaluate to what extent violations of the assumptions in our study  impact the performance of causal methods and to what extent assumptions such as overlap and unconfoundedness are violated in practice, potentially requiring the development of alternative solutions. Such research might have critical importance to the adoption of causal machine learning in practice. We link this stream of future research to \citet{Bertsimas.2022} and the notion that for operational decision-making, a true, unbiased causal model (by the standards of meeting all requirements discussed in Section~\ref{sec:Method_Assumptions}) might not always be needed. Third, we highlight the need for research in identifying the strength of selection bias in data. While counterfactual data will never be available for modeling, the strength of selection bias could be assessed. Such a methodology would be able to inform decision makers on whether a causal approach must be considered. Finally, our performance metrics rely on relative performance per customer and do not take absolute costs and revenues into account. We believe that pricing would therefore be a promising application of cost-sensitive approaches in causal inference, such as introduced in \citet{Verbeke.2020} and \citet{Verbeke.2023}.

\section*{Acknowledgements\label{sec:Acknowledgements}}

\noindent The authors want to thank AXA Bank Belgium, Crelan, and the involved project team who have supported this project throughout by providing feedback, know-how, and access to the data used for this research project.\\

\noindent This work was supported by the FWO research project G015020N.

\section*{Conflict of interest\label{sec:CoI}}

\noindent The authors state no conflict of interest.

\bibliographystyle{vancouver}
\bibliography{main}

\begin{thebibliography}{65}
\expandafter\ifx\csname natexlab\endcsname\relax\def\natexlab#1{#1}\fi
\providecommand{\url}[1]{\texttt{#1}}
\providecommand{\href}[2]{#2}
\providecommand{\path}[1]{#1}
\providecommand{\DOIprefix}{doi:}
\providecommand{\ArXivprefix}{arXiv:}
\providecommand{\URLprefix}{URL: }
\providecommand{\Pubmedprefix}{pmid:}
\providecommand{\doi}[1]{\href{http://dx.doi.org/#1}{\path{#1}}}
\providecommand{\Pubmed}[1]{\href{pmid:#1}{\path{#1}}}
\providecommand{\bibinfo}[2]{#2}
\ifx\xfnm\relax \def\xfnm[#1]{\unskip,\space#1}\fi
\bibitem[{Adam et~al.(2020)Adam, Chang, Haibe-Kains and Goldenberg}]{Adam.2020}
\bibinfo{author}{Adam, G.A.}, \bibinfo{author}{Chang, C.H.K.},
  \bibinfo{author}{Haibe-Kains, B.}, \bibinfo{author}{Goldenberg, A.},
  \bibinfo{year}{2020}.
\newblock \bibinfo{title}{Hidden risks of machine learning applied to
  healthcare: unintended feedback loops between models and future data causing
  model degradation}, in: \bibinfo{booktitle}{Machine Learning for Healthcare
  Conference}, \bibinfo{organization}{PMLR}. pp. \bibinfo{pages}{710--731}.
\bibitem[{Agrawal and Ferguson(2007)}]{Agrawal.2007}
\bibinfo{author}{Agrawal, V.}, \bibinfo{author}{Ferguson, M.},
  \bibinfo{year}{2007}.
\newblock \bibinfo{title}{Bid-response models for customised pricing}.
\newblock \bibinfo{journal}{Journal of Revenue and Pricing Management}
  \bibinfo{volume}{6}, \bibinfo{pages}{212--228}.
\bibitem[{Angrist and Pischke(2009)}]{Angrist.2009}
\bibinfo{author}{Angrist, J.D.}, \bibinfo{author}{Pischke, J.S.},
  \bibinfo{year}{2009}.
\newblock \bibinfo{title}{Mostly harmless econometrics: An empiricist's
  companion}.
\newblock \bibinfo{publisher}{Princeton university press}.
\bibitem[{Arevalillo(2019)}]{Arevalillo.2019}
\bibinfo{author}{Arevalillo, J.M.}, \bibinfo{year}{2019}.
\newblock \bibinfo{title}{A machine learning approach to assess price
  sensitivity with application to automobile loan segmentation}.
\newblock \bibinfo{journal}{Applied Soft Computing} \bibinfo{volume}{76},
  \bibinfo{pages}{390--399}.
\newblock \URLprefix
  \url{https://www.sciencedirect.com/science/article/pii/S1568494618307014},
  \DOIprefix\doi{https://doi.org/10.1016/j.asoc.2018.12.012}.
\bibitem[{Berhold(1971)}]{Berhold.1971}
\bibinfo{author}{Berhold, M.}, \bibinfo{year}{1971}.
\newblock \bibinfo{title}{A theory of linear profit-sharing incentives}.
\newblock \bibinfo{journal}{The Quarterly Journal of Economics}
  \bibinfo{volume}{85}, \bibinfo{pages}{460--482}.
\bibitem[{Berrevoets et~al.(2020)Berrevoets, Jordon, Bica, van~der Schaar
  et~al.}]{Berrevoets.2020}
\bibinfo{author}{Berrevoets, J.}, \bibinfo{author}{Jordon, J.},
  \bibinfo{author}{Bica, I.}, \bibinfo{author}{van~der Schaar, M.}, et~al.,
  \bibinfo{year}{2020}.
\newblock \bibinfo{title}{Organite: Optimal transplant donor organ offering
  using an individual treatment effect}.
\newblock \bibinfo{journal}{Advances in neural information processing systems}
  \bibinfo{volume}{33}, \bibinfo{pages}{20037--20050}.
\bibitem[{Berry(1994)}]{Berry.1994}
\bibinfo{author}{Berry, S.T.}, \bibinfo{year}{1994}.
\newblock \bibinfo{title}{Estimating discrete-choice models of product
  differentiation}.
\newblock \bibinfo{journal}{The RAND Journal of Economics} ,
  \bibinfo{pages}{242--262}.
\bibitem[{Berry et~al.(1995)Berry, Levinsohn and Pakes}]{Berry.1995}
\bibinfo{author}{Berry, S.T.}, \bibinfo{author}{Levinsohn, J.},
  \bibinfo{author}{Pakes, A.}, \bibinfo{year}{1995}.
\newblock \bibinfo{title}{Automobile prices in market equilibrium}.
\newblock \bibinfo{journal}{Econometrica} \bibinfo{volume}{63},
  \bibinfo{pages}{841--890}.
\newblock \URLprefix \url{http://www.jstor.org/stable/2171802}.
\bibitem[{Bertsimas and Kallus(2022)}]{Bertsimas.2022}
\bibinfo{author}{Bertsimas, D.}, \bibinfo{author}{Kallus, N.},
  \bibinfo{year}{2022}.
\newblock \bibinfo{title}{{The Power and Limits of Predictive Approaches to
  Observational Data-Driven Optimization: The Case of Pricing}}.
\newblock \bibinfo{journal}{{INFORMS Journal on Optimization}}
  \DOIprefix\doi{10.1287/ijoo.2022.0077}.
\bibitem[{Bica et~al.(2020)Bica, Jordon and van~der Schaar}]{Bica.2020}
\bibinfo{author}{Bica, I.}, \bibinfo{author}{Jordon, J.},
  \bibinfo{author}{van~der Schaar, M.}, \bibinfo{year}{2020}.
\newblock \bibinfo{title}{Estimating the effects of continuous-valued
  interventions using generative adversarial networks}.
\newblock \bibinfo{journal}{Advances in Neural Information Processing Systems}
  \bibinfo{volume}{33}, \bibinfo{pages}{16434--16445}.
\bibitem[{Breiman(2001)}]{Breiman.2001}
\bibinfo{author}{Breiman, L.}, \bibinfo{year}{2001}.
\newblock \bibinfo{title}{Random forests}.
\newblock \bibinfo{journal}{Machine learning} \bibinfo{volume}{45},
  \bibinfo{pages}{5--32}.
\bibitem[{Brier et~al.(1950)}]{Brier.1950}
\bibinfo{author}{Brier, G.W.}, et~al., \bibinfo{year}{1950}.
\newblock \bibinfo{title}{Verification of forecasts expressed in terms of
  probability}.
\newblock \bibinfo{journal}{Monthly weather review} \bibinfo{volume}{78},
  \bibinfo{pages}{1--3}.
\bibitem[{Cozarenco and Szafarz(2018)}]{Cozarenco.2018}
\bibinfo{author}{Cozarenco, A.}, \bibinfo{author}{Szafarz, A.},
  \bibinfo{year}{2018}.
\newblock \bibinfo{title}{Gender biases in bank lending: Lessons from
  microcredit in france}.
\newblock \bibinfo{journal}{Journal of Business Ethics} \bibinfo{volume}{147},
  \bibinfo{pages}{631--650}.
\bibitem[{Cramer(2002)}]{Cramer.2002}
\bibinfo{author}{Cramer, J.S.}, \bibinfo{year}{2002}.
\newblock \bibinfo{title}{The origins of logistic regression} .
\bibitem[{Deaton and Cartwright(2018)}]{Deaton.2018}
\bibinfo{author}{Deaton, A.}, \bibinfo{author}{Cartwright, N.},
  \bibinfo{year}{2018}.
\newblock \bibinfo{title}{Understanding and misunderstanding randomized
  controlled trials}.
\newblock \bibinfo{journal}{Social science \& medicine} \bibinfo{volume}{210},
  \bibinfo{pages}{2--21}.
\bibitem[{Ferkingstad et~al.(2011)Ferkingstad, L{\o}land and
  Wilhelmsen}]{Ferkingstad.2011}
\bibinfo{author}{Ferkingstad, E.}, \bibinfo{author}{L{\o}land, A.},
  \bibinfo{author}{Wilhelmsen, M.}, \bibinfo{year}{2011}.
\newblock \bibinfo{title}{Causal modeling and inference for electricity
  markets}.
\newblock \bibinfo{journal}{Energy Economics} \bibinfo{volume}{33},
  \bibinfo{pages}{404--412}.
\bibitem[{Garrow et~al.(2006)Garrow, Ferguson, Keskinocak and
  Swann}]{Garrow.2006}
\bibinfo{author}{Garrow, L.}, \bibinfo{author}{Ferguson, M.},
  \bibinfo{author}{Keskinocak, P.}, \bibinfo{author}{Swann, J.},
  \bibinfo{year}{2006}.
\newblock \bibinfo{title}{Expert opinions: Current pricing and revenue
  management practice across us industries}.
\newblock \bibinfo{journal}{Journal of revenue and pricing management}
  \bibinfo{volume}{5}, \bibinfo{pages}{237--247}.
\bibitem[{Guelman and Guill{\'e}n(2014)}]{Guelman.2014}
\bibinfo{author}{Guelman, L.}, \bibinfo{author}{Guill{\'e}n, M.},
  \bibinfo{year}{2014}.
\newblock \bibinfo{title}{A causal inference approach to measure price
  elasticity in automobile insurance}.
\newblock \bibinfo{journal}{Expert Systems with Applications}
  \bibinfo{volume}{41}, \bibinfo{pages}{387--396}.
\bibitem[{Hanssens et~al.(2003)Hanssens, Parsons and Schultz}]{Hanssens.2003}
\bibinfo{author}{Hanssens, D.M.}, \bibinfo{author}{Parsons, L.J.},
  \bibinfo{author}{Schultz, R.L.}, \bibinfo{year}{2003}.
\newblock \bibinfo{title}{Market response models: Econometric and time series
  analysis}. volume~\bibinfo{volume}{2}.
\newblock \bibinfo{publisher}{Springer Science \& Business Media}.
\bibitem[{Hausman et~al.(2012)Hausman, Newey, Woutersen, Chao and
  Swanson}]{Hausman.2012}
\bibinfo{author}{Hausman, J.A.}, \bibinfo{author}{Newey, W.K.},
  \bibinfo{author}{Woutersen, T.}, \bibinfo{author}{Chao, J.C.},
  \bibinfo{author}{Swanson, N.R.}, \bibinfo{year}{2012}.
\newblock \bibinfo{title}{Instrumental variable estimation with
  heteroskedasticity and many instruments}.
\newblock \bibinfo{journal}{Quantitative Economics} \bibinfo{volume}{3},
  \bibinfo{pages}{211--255}.
\bibitem[{Heckman(1978)}]{Heckman.1978}
\bibinfo{author}{Heckman, J.J.}, \bibinfo{year}{1978}.
\newblock \bibinfo{title}{Simple statistical models for discrete panel data
  developed and applied to test the hypothesis of true state dependence against
  the hypothesis of spurious state dependence}, in: \bibinfo{booktitle}{Annales
  de l'INSEE}, \bibinfo{organization}{JSTOR}. pp. \bibinfo{pages}{227--269}.
\bibitem[{Heckman(1990)}]{Heckman.1990}
\bibinfo{author}{Heckman, J.J.}, \bibinfo{year}{1990}.
\newblock \bibinfo{title}{Selection bias and self-selection}.
\newblock \bibinfo{journal}{Econometrics} , \bibinfo{pages}{201--224}.
\bibitem[{Hirano and Imbens(2004)}]{Hirano.2004}
\bibinfo{author}{Hirano, K.}, \bibinfo{author}{Imbens, G.W.},
  \bibinfo{year}{2004}.
\newblock \bibinfo{title}{The propensity score with continuous treatments}.
\newblock \bibinfo{journal}{Applied Bayesian modeling and causal inference from
  incomplete-data perspectives} \bibinfo{volume}{226164},
  \bibinfo{pages}{73--84}.
\bibitem[{Holland(1986)}]{Holland.1986}
\bibinfo{author}{Holland, P.W.}, \bibinfo{year}{1986}.
\newblock \bibinfo{title}{Statistics and causal inference}.
\newblock \bibinfo{journal}{Journal of the American statistical Association}
  \bibinfo{volume}{81}, \bibinfo{pages}{945--960}.
\bibitem[{Hornik et~al.(1989)Hornik, Stinchcombe and White}]{Hornik.1989}
\bibinfo{author}{Hornik, K.}, \bibinfo{author}{Stinchcombe, M.},
  \bibinfo{author}{White, H.}, \bibinfo{year}{1989}.
\newblock \bibinfo{title}{Multilayer feedforward networks are universal
  approximators}.
\newblock \bibinfo{journal}{Neural networks} \bibinfo{volume}{2},
  \bibinfo{pages}{359--366}.
\bibitem[{Imai and Van~Dyk(2004)}]{Imai.2004}
\bibinfo{author}{Imai, K.}, \bibinfo{author}{Van~Dyk, D.A.},
  \bibinfo{year}{2004}.
\newblock \bibinfo{title}{Causal inference with general treatment regimes:
  Generalizing the propensity score}.
\newblock \bibinfo{journal}{Journal of the American Statistical Association}
  \bibinfo{volume}{99}, \bibinfo{pages}{854--866}.
\bibitem[{Imbens(2000)}]{Imbens.2000}
\bibinfo{author}{Imbens, G.W.}, \bibinfo{year}{2000}.
\newblock \bibinfo{title}{The role of the propensity score in estimating
  dose-response functions}.
\newblock \bibinfo{journal}{Biometrika} \bibinfo{volume}{87},
  \bibinfo{pages}{706--710}.
\bibitem[{Jain et~al.(2016)Jain, Gyanchandani and Khare}]{Jain.2016}
\bibinfo{author}{Jain, P.}, \bibinfo{author}{Gyanchandani, M.},
  \bibinfo{author}{Khare, N.}, \bibinfo{year}{2016}.
\newblock \bibinfo{title}{Big data privacy: a technological perspective and
  review}.
\newblock \bibinfo{journal}{Journal of Big Data} \bibinfo{volume}{3},
  \bibinfo{pages}{1--25}.
\bibitem[{Kuksov and Villas-Boas(2008)}]{Kuksov.2008}
\bibinfo{author}{Kuksov, D.}, \bibinfo{author}{Villas-Boas, J.M.},
  \bibinfo{year}{2008}.
\newblock \bibinfo{title}{Endogeneity and individual consumer choice}.
\newblock \bibinfo{journal}{Journal of Marketing Research}
  \bibinfo{volume}{45}, \bibinfo{pages}{702--714}.
\bibitem[{Lawrence(2003)}]{Lawrence.2003}
\bibinfo{author}{Lawrence, R.D.}, \bibinfo{year}{2003}.
\newblock \bibinfo{title}{A machine-learning approach to optimal bid pricing},
  in: \bibinfo{booktitle}{Computational modeling and problem solving in the
  networked world}. \bibinfo{publisher}{Springer}, pp.
  \bibinfo{pages}{97--118}.
\bibitem[{Lechner(2001)}]{Lechner.2001}
\bibinfo{author}{Lechner, M.}, \bibinfo{year}{2001}.
\newblock \bibinfo{title}{Identification and estimation of causal effects of
  multiple treatments under the conditional independence assumption}.
\newblock \bibinfo{publisher}{Springer}.
\bibitem[{Mansoury et~al.(2020)Mansoury, Abdollahpouri, Pechenizkiy, Mobasher
  and Burke}]{Mansoury.2020}
\bibinfo{author}{Mansoury, M.}, \bibinfo{author}{Abdollahpouri, H.},
  \bibinfo{author}{Pechenizkiy, M.}, \bibinfo{author}{Mobasher, B.},
  \bibinfo{author}{Burke, R.}, \bibinfo{year}{2020}.
\newblock \bibinfo{title}{Feedback loop and bias amplification in recommender
  systems}, in: \bibinfo{booktitle}{Proceedings of the 29th ACM international
  conference on information \& knowledge management}, pp.
  \bibinfo{pages}{2145--2148}.
\bibitem[{Newey(1987)}]{Newey.1987}
\bibinfo{author}{Newey, W.K.}, \bibinfo{year}{1987}.
\newblock \bibinfo{title}{Efficient estimation of limited dependent variable
  models with endogenous explanatory variables}.
\newblock \bibinfo{journal}{Journal of econometrics} \bibinfo{volume}{36},
  \bibinfo{pages}{231--250}.
\bibitem[{Nie et~al.(2021)Nie, Ye, Liu and Nicolae}]{Nie.2021}
\bibinfo{author}{Nie, L.}, \bibinfo{author}{Ye, M.}, \bibinfo{author}{Liu, Q.},
  \bibinfo{author}{Nicolae, D.}, \bibinfo{year}{2021}.
\newblock \bibinfo{title}{Vcnet and functional targeted regularization for
  learning causal effects of continuous treatments}.
\newblock \bibinfo{journal}{arXiv preprint arXiv:2103.07861} .
\bibitem[{Oliver and Oliver(2014)}]{Oliver.2014}
\bibinfo{author}{Oliver, B.V.}, \bibinfo{author}{Oliver, R.M.},
  \bibinfo{year}{2014}.
\newblock \bibinfo{title}{Optimal roe loan pricing with or without adverse
  selection}.
\newblock \bibinfo{journal}{Journal of the Operational Research Society}
  \bibinfo{volume}{65}, \bibinfo{pages}{435--442}.
\bibitem[{Paszke et~al.(2017)Paszke, Gross, Chintala, Chanan, Yang, DeVito,
  Lin, Desmaison, Antiga and Lerer}]{Paszke.2017}
\bibinfo{author}{Paszke, A.}, \bibinfo{author}{Gross, S.},
  \bibinfo{author}{Chintala, S.}, \bibinfo{author}{Chanan, G.},
  \bibinfo{author}{Yang, E.}, \bibinfo{author}{DeVito, Z.},
  \bibinfo{author}{Lin, Z.}, \bibinfo{author}{Desmaison, A.},
  \bibinfo{author}{Antiga, L.}, \bibinfo{author}{Lerer, A.},
  \bibinfo{year}{2017}.
\newblock \bibinfo{title}{Automatic differentiation in pytorch}, in:
  \bibinfo{booktitle}{NIPS-W}.
\bibitem[{Pedregosa et~al.(2011)Pedregosa, Varoquaux, Gramfort, Michel,
  Thirion, Grisel, Blondel, Prettenhofer, Weiss, Dubourg, Vanderplas, Passos,
  Cournapeau, Brucher, Perrot and Duchesnay}]{Scikit.2011}
\bibinfo{author}{Pedregosa, F.}, \bibinfo{author}{Varoquaux, G.},
  \bibinfo{author}{Gramfort, A.}, \bibinfo{author}{Michel, V.},
  \bibinfo{author}{Thirion, B.}, \bibinfo{author}{Grisel, O.},
  \bibinfo{author}{Blondel, M.}, \bibinfo{author}{Prettenhofer, P.},
  \bibinfo{author}{Weiss, R.}, \bibinfo{author}{Dubourg, V.},
  \bibinfo{author}{Vanderplas, J.}, \bibinfo{author}{Passos, A.},
  \bibinfo{author}{Cournapeau, D.}, \bibinfo{author}{Brucher, M.},
  \bibinfo{author}{Perrot, M.}, \bibinfo{author}{Duchesnay, E.},
  \bibinfo{year}{2011}.
\newblock \bibinfo{title}{Scikit-learn: Machine learning in {P}ython}.
\newblock \bibinfo{journal}{Journal of Machine Learning Research}
  \bibinfo{volume}{12}, \bibinfo{pages}{2825--2830}.
\bibitem[{Phillips(2012)}]{Phillips.2012}
\bibinfo{author}{Phillips, R.}, \bibinfo{year}{2012}.
\newblock \bibinfo{title}{{Customized Pricing}}, in: \bibinfo{booktitle}{{The
  Oxford Handbook of Pricing Management}}. \bibinfo{publisher}{Oxford
  University Press}.
\newblock \URLprefix
  \url{https://doi.org/10.1093/oxfordhb/9780199543175.013.0021},
  \DOIprefix\doi{10.1093/oxfordhb/9780199543175.013.0021}.
\bibitem[{Phillips(2013)}]{Phillips.2013}
\bibinfo{author}{Phillips, R.}, \bibinfo{year}{2013}.
\newblock \bibinfo{title}{Optimizing prices for consumer credit}.
\newblock \bibinfo{journal}{Journal of Revenue and Pricing Management}
  \bibinfo{volume}{12}, \bibinfo{pages}{360--377}.
\bibitem[{Phillips(2021)}]{Phillips.2021}
\bibinfo{author}{Phillips, R.}, \bibinfo{year}{2021}.
\newblock \bibinfo{title}{{Pricing and revenue optimization}}.
\newblock \bibinfo{edition}{Second edition} ed., \bibinfo{publisher}{{Stanford
  Business Books}}, \bibinfo{address}{Stanford, California}.
\bibitem[{Phillips et~al.(2012)Phillips, Simsek and Van~Ryzin}]{Phillips.2012a}
\bibinfo{author}{Phillips, R.}, \bibinfo{author}{Simsek, A.S.},
  \bibinfo{author}{Van~Ryzin, G.}, \bibinfo{year}{2012}.
\newblock \bibinfo{title}{Endogeneity and price sensitivity in customized
  pricing}.
\newblock \bibinfo{journal}{Columbia University Center for Pricing and Revenue
  Management Working Paper} \bibinfo{volume}{4}.
\bibitem[{Phillips et~al.(2015)Phillips, {\c{S}}im{\c{s}}ek and
  Van~Ryzin}]{Phillips.2015}
\bibinfo{author}{Phillips, R.}, \bibinfo{author}{{\c{S}}im{\c{s}}ek, A.S.},
  \bibinfo{author}{Van~Ryzin, G.}, \bibinfo{year}{2015}.
\newblock \bibinfo{title}{The effectiveness of field price discretion:
  Empirical evidence from auto lending}.
\newblock \bibinfo{journal}{Management Science} \bibinfo{volume}{61},
  \bibinfo{pages}{1741--1759}.
\bibitem[{Qian et~al.(2023)Qian, Cebere and van~der Schaar}]{Qian.2023}
\bibinfo{author}{Qian, Z.}, \bibinfo{author}{Cebere, B.C.},
  \bibinfo{author}{van~der Schaar, M.}, \bibinfo{year}{2023}.
\newblock \bibinfo{title}{Synthcity: facilitating innovative use cases of
  synthetic data in different data modalities}.
\newblock \bibinfo{journal}{arXiv preprint arXiv:2301.07573} .
\bibitem[{Radcliffe(2007)}]{Radcliffe.2007}
\bibinfo{author}{Radcliffe, N.}, \bibinfo{year}{2007}.
\newblock \bibinfo{title}{Using control groups to target on predicted lift:
  Building and assessing uplift model}.
\newblock \bibinfo{journal}{Direct Marketing Analytics Journal} ,
  \bibinfo{pages}{14--21}.
\bibitem[{Richards(1959)}]{Richards.1959}
\bibinfo{author}{Richards, F.}, \bibinfo{year}{1959}.
\newblock \bibinfo{title}{A flexible growth function for empirical use}.
\newblock \bibinfo{journal}{Journal of experimental Botany}
  \bibinfo{volume}{10}, \bibinfo{pages}{290--301}.
\bibitem[{Rivers and Vuong(1988)}]{Rivers.1988}
\bibinfo{author}{Rivers, D.}, \bibinfo{author}{Vuong, Q.H.},
  \bibinfo{year}{1988}.
\newblock \bibinfo{title}{Limited information estimators and exogeneity tests
  for simultaneous probit models}.
\newblock \bibinfo{journal}{Journal of econometrics} \bibinfo{volume}{39},
  \bibinfo{pages}{347--366}.
\bibitem[{Rosenbaum and Rubin(1983)}]{Rosenbaum.1983}
\bibinfo{author}{Rosenbaum, P.R.}, \bibinfo{author}{Rubin, D.B.},
  \bibinfo{year}{1983}.
\newblock \bibinfo{title}{The central role of the propensity score in
  observational studies for causal effects}.
\newblock \bibinfo{journal}{Biometrika} \bibinfo{volume}{70},
  \bibinfo{pages}{41--55}.
\bibitem[{Rosenbaum and Rubin(1984)}]{Rosenbaum.1984}
\bibinfo{author}{Rosenbaum, P.R.}, \bibinfo{author}{Rubin, D.B.},
  \bibinfo{year}{1984}.
\newblock \bibinfo{title}{Reducing bias in observational studies using
  subclassification on the propensity score}.
\newblock \bibinfo{journal}{Journal of the American statistical Association}
  \bibinfo{volume}{79}, \bibinfo{pages}{516--524}.
\bibitem[{Ross(1973)}]{Ross.1973}
\bibinfo{author}{Ross, S.A.}, \bibinfo{year}{1973}.
\newblock \bibinfo{title}{The economic theory of agency: The principal's
  problem}.
\newblock \bibinfo{journal}{The American economic review} \bibinfo{volume}{63},
  \bibinfo{pages}{134--139}.
\bibitem[{Rubin(1974)}]{Rubin.1974}
\bibinfo{author}{Rubin, D.B.}, \bibinfo{year}{1974}.
\newblock \bibinfo{title}{Estimating causal effects of treatments in randomized
  and nonrandomized studies}.
\newblock \bibinfo{journal}{Journal of educational Psychology}
  \bibinfo{volume}{66}, \bibinfo{pages}{688}.
\bibitem[{Rubin(2004)}]{Rubin.2004}
\bibinfo{author}{Rubin, D.B.}, \bibinfo{year}{2004}.
\newblock \bibinfo{title}{Direct and indirect causal effects via potential
  outcomes}.
\newblock \bibinfo{journal}{Scandinavian Journal of Statistics}
  \bibinfo{volume}{31}, \bibinfo{pages}{161--170}.
\bibitem[{Rubin(2005)}]{Rubin.2005}
\bibinfo{author}{Rubin, D.B.}, \bibinfo{year}{2005}.
\newblock \bibinfo{title}{Causal inference using potential outcomes: Design,
  modeling, decisions}.
\newblock \bibinfo{journal}{Journal of the American Statistical Association}
  \bibinfo{volume}{100}, \bibinfo{pages}{322--331}.
\bibitem[{van Ryzin(2012)}]{vanRyzin.2012}
\bibinfo{author}{van Ryzin, G.J.}, \bibinfo{year}{2012}.
\newblock \bibinfo{title}{{Models of Demand}}, in: \bibinfo{booktitle}{{The
  Oxford Handbook of Pricing Management}}. \bibinfo{publisher}{Oxford
  University Press}.
\newblock \URLprefix
  \url{https://doi.org/10.1093/oxfordhb/9780199543175.013.0018},
  \DOIprefix\doi{10.1093/oxfordhb/9780199543175.013.0018}.
\bibitem[{Schwab et~al.(2020)Schwab, Linhardt, Bauer, Buhmann and
  Karlen}]{Schwab.2020}
\bibinfo{author}{Schwab, P.}, \bibinfo{author}{Linhardt, L.},
  \bibinfo{author}{Bauer, S.}, \bibinfo{author}{Buhmann, J.M.},
  \bibinfo{author}{Karlen, W.}, \bibinfo{year}{2020}.
\newblock \bibinfo{title}{Learning counterfactual representations for
  estimating individual dose-response curves}, in:
  \bibinfo{booktitle}{Proceedings of the AAAI Conference on Artificial
  Intelligence}, pp. \bibinfo{pages}{5612--5619}.
\bibitem[{Seabold and Perktold(2010)}]{Seabold.2010}
\bibinfo{author}{Seabold, S.}, \bibinfo{author}{Perktold, J.},
  \bibinfo{year}{2010}.
\newblock \bibinfo{title}{statsmodels: Econometric and statistical modeling
  with python}, in: \bibinfo{booktitle}{9th Python in Science Conference}.
\bibitem[{Shalit et~al.(2017)Shalit, Johansson and Sontag}]{Shalit.2017}
\bibinfo{author}{Shalit, U.}, \bibinfo{author}{Johansson, F.D.},
  \bibinfo{author}{Sontag, D.}, \bibinfo{year}{2017}.
\newblock \bibinfo{title}{Estimating individual treatment effect:
  generalization bounds and algorithms}, in: \bibinfo{booktitle}{International
  Conference on Machine Learning}, \bibinfo{organization}{PMLR}. pp.
  \bibinfo{pages}{3076--3085}.
\bibitem[{Sundararajan et~al.(2011)Sundararajan, Bhaskar, Sarkar, Dasaratha,
  Bal, Marasanapalle, Zmudzka and Bak}]{Sundararajan.2011}
\bibinfo{author}{Sundararajan, R.}, \bibinfo{author}{Bhaskar, T.},
  \bibinfo{author}{Sarkar, A.}, \bibinfo{author}{Dasaratha, S.},
  \bibinfo{author}{Bal, D.}, \bibinfo{author}{Marasanapalle, J.K.},
  \bibinfo{author}{Zmudzka, B.}, \bibinfo{author}{Bak, K.},
  \bibinfo{year}{2011}.
\newblock \bibinfo{title}{Marketing optimization in retail banking}.
\newblock \bibinfo{journal}{Interfaces} \bibinfo{volume}{41},
  \bibinfo{pages}{485--505}.
\bibitem[{Vanderschueren et~al.(2023)Vanderschueren, Boute, Verdonck, Baesens
  and Verbeke}]{Vanderschueren.2022}
\bibinfo{author}{Vanderschueren, T.}, \bibinfo{author}{Boute, R.},
  \bibinfo{author}{Verdonck, T.}, \bibinfo{author}{Baesens, B.},
  \bibinfo{author}{Verbeke, W.}, \bibinfo{year}{2023}.
\newblock \bibinfo{title}{Optimizing the preventive maintenance frequency with
  causal machine learning}.
\newblock \bibinfo{journal}{International Journal of Production Economics}
  \bibinfo{volume}{258}, \bibinfo{pages}{108798}.
\bibitem[{Varian(1989)}]{Varian.1989}
\bibinfo{author}{Varian, H.R.}, \bibinfo{year}{1989}.
\newblock \bibinfo{title}{Price discrimination}.
\newblock \bibinfo{journal}{Handbook of industrial organization}
  \bibinfo{volume}{1}, \bibinfo{pages}{597--654}.
\bibitem[{Varian(2014)}]{Varian.2014}
\bibinfo{author}{Varian, H.R.}, \bibinfo{year}{2014}.
\newblock \bibinfo{title}{Intermediate microeconomics with calculus: a modern
  approach}.
\newblock \bibinfo{publisher}{WW norton \& company}.
\bibitem[{Varian(2016)}]{Varian.2016}
\bibinfo{author}{Varian, H.R.}, \bibinfo{year}{2016}.
\newblock \bibinfo{title}{Causal inference in economics and marketing}.
\newblock \bibinfo{journal}{Proceedings of the National Academy of Sciences}
  \bibinfo{volume}{113}, \bibinfo{pages}{7310--7315}.
\bibitem[{Verbeke et~al.(2020)Verbeke, Olaya, Berrevoets, Verboven and
  Maldonado}]{Verbeke.2020}
\bibinfo{author}{Verbeke, W.}, \bibinfo{author}{Olaya, D.},
  \bibinfo{author}{Berrevoets, J.}, \bibinfo{author}{Verboven, S.},
  \bibinfo{author}{Maldonado, S.}, \bibinfo{year}{2020}.
\newblock \bibinfo{title}{The foundations of cost-sensitive causal
  classification}.
\newblock \bibinfo{journal}{10.48550/arXiv.2007.12582} .
\bibitem[{Verbeke et~al.(2023)Verbeke, Olaya, Guerry and
  Van~Belle}]{Verbeke.2023}
\bibinfo{author}{Verbeke, W.}, \bibinfo{author}{Olaya, D.},
  \bibinfo{author}{Guerry, M.A.}, \bibinfo{author}{Van~Belle, J.},
  \bibinfo{year}{2023}.
\newblock \bibinfo{title}{To do or not to do? cost-sensitive causal
  classification with individual treatment effect estimates}.
\newblock \bibinfo{journal}{European Journal of Operational Research}
  \bibinfo{volume}{305}, \bibinfo{pages}{838--852}.
\bibitem[{Villas-Boas and Winer(1999)}]{Villas.1999}
\bibinfo{author}{Villas-Boas, J.M.}, \bibinfo{author}{Winer, R.S.},
  \bibinfo{year}{1999}.
\newblock \bibinfo{title}{Endogeneity in brand choice models}.
\newblock \bibinfo{journal}{Management science} \bibinfo{volume}{45},
  \bibinfo{pages}{1324--1338}.
\bibitem[{Yoon et~al.(2018)Yoon, Jordon and Van Der~Schaar}]{Yoon.2018}
\bibinfo{author}{Yoon, J.}, \bibinfo{author}{Jordon, J.}, \bibinfo{author}{Van
  Der~Schaar, M.}, \bibinfo{year}{2018}.
\newblock \bibinfo{title}{Ganite: Estimation of individualized treatment
  effects using generative adversarial nets}, in:
  \bibinfo{booktitle}{International Conference on Learning Representations}.
\newblock \URLprefix \url{https://openreview.net/forum?id=ByKWUeWA-}.

\end{thebibliography}

\setcounter{table}{0}
\renewcommand{\thetable}{A\arabic{table}}
\setcounter{figure}{0}
\renewcommand{\thefigure}{A\arabic{figure}}

\begin{appendix}
\section{Experimental results\label{sec:Append_Results}}

\begin{table}[h]
    \centering
    \caption{Policy Error (PE) on Richards curve}
    \footnotesize
    \begin{tabular}{lccccccc:c}
        \multicolumn{9}{l}{\textbf{Ground truth:} Richards curve}\\
            \hline\\[-2.3ex]
	    \hline &&&&&&&&\\[-2.0ex] 
	    & \multicolumn{7}{c:}{\textit{Bias strength}} & \multirow{2}{*}{True bids}\\
	    \cdashline{2-8}&&&&&&&&\\[-2.0ex]
        Model & 0.0 & 2.5 & 5.0 & 7.5 & 10.0 & 15.0 & 20.0 & \\
        \hline &&&&&&&&\\[-1.0ex]
        Na\"ive pricing & 	0.307 & 	0.212 & 	0.190 & 	0.182 & 	0.176 & 	0.171 & 	0.170 & 	0.228	\\
        Logistic Regression & 	\textit{0.049} & 	0.044 & 	\textit{0.043} & 	\textit{0.043} & 	\textit{0.046} & 	\textit{0.047} & 	\textit{0.051} & 	\textit{0.057}	\\
        Random Forest & 	0.083 & 	0.121 & 	0.166 & 	0.188 & 	0.208 & 	0.221 & 	0.258 & 	0.281	\\
        MLP & 	\textbf{0.023} & 	\textbf{0.022} & 	\textbf{0.018} & 	\textbf{0.020} & 	\textbf{0.022} & 	\textbf{0.018} & 	\textbf{0.022} & 	\textbf{0.024}	\\
        \hdashline&&&&&&&&\\[-2.0ex]									
        HIE & 	0.074 & 	0.071 & 	0.071 & 	0.070 & 	0.071 & 	0.071 & 	0.076 & 	0.076	\\
        DRNets & 	0.051 & 	0.071 & 	0.160 & 	0.167 & 	0.096 & 	0.093 & 	0.156 & 	0.176	\\
        VCNets & 	\textit{0.049} & 	\textit{0.043} & 	\textit{0.043} & 	0.050 & 	0.056 & 	0.059 & 	0.065 & 	0.060	\\
        \hline \\[-1.0ex]
    \end{tabular}
    \label{tbl:Results_PEs}
\end{table}

\begin{table}[h]
    \centering
    \caption{Policy Error (PE) on stacked sigmoid}
    \footnotesize
    \begin{tabular}{lccccccc:c}
        \multicolumn{9}{l}{\textbf{Ground truth:} Stacked sigmoid}\\
            \hline\\[-2.3ex]
	    \hline &&&&&&&&\\[-2.0ex] 
	    & \multicolumn{7}{c:}{\textit{Bias strength}} & \multirow{2}{*}{True bids}\\
	    \cdashline{2-8}&&&&&&&&\\[-2.0ex]
        Model & 0.0 & 2.5 & 5.0 & 7.5 & 10.0 & 15.0 & 20.0 & \\
        \hline &&&&&&&&\\[-1.0ex]
        Na\"ive pricing & 	0.350 & 	0.294 & 	0.285 & 	0.289 & 	0.279 & 	0.291 & 	0.292 & 	0.417	\\
        Logistic Regression & 	0.245 & 	0.173 & 	0.150 & 	0.153 & 	0.154 & 	\textit{0.162} & 	\textit{0.145} & 	0.270	\\
        Random Forest & 	\textit{0.090} & 	0.142 & 	0.200 & 	0.238 & 	0.249 & 	0.256 & 	0.282 & 	0.325	\\
        MLP & 	\textbf{0.079} & 	0.118 & 	0.139 & 	0.170 & 	0.203 & 	0.205 & 	0.220 & 	0.235	\\
        \hdashline&&&&&&&&\\[-2.0ex]									
        HIE & 	0.092 & 	0.121 & 	0.143 & 	0.167 & 	0.187 & 	0.295 & 	0.276 & 	0.324	\\
        DRNets & 	\textit{0.090} & 	\textit{0.109} & 	\textit{0.127} & 	\textbf{0.109} & 	\textbf{0.122} & 	\textbf{0.122} & 	\textbf{0.142} & 	\textbf{0.137}	\\
        VCNets & 	0.132 & 	\textbf{0.093} & 	\textbf{0.124} & 	\textit{0.129} & 	\textit{0.130} & 	0.192 & 	0.157 & 	\textit{0.151}	\\
        \hline \\[-1.0ex]
    \end{tabular}
    \label{tbl:Results_PEs_SS}
\end{table}

\begin{table}[ht]
    \centering
    \caption{Brier score (BS) on stacked sigmoid}
    \footnotesize
    \begin{tabular}{lccccccc:c}
        \multicolumn{9}{l}{\textbf{Ground truth:} Stacked sigmoid}\\
            \hline\\[-2.3ex]
	    \hline &&&&&&&&\\[-2.0ex] 
	    & \multicolumn{7}{c:}{\textit{Bias strength}} & \multirow{2}{*}{True bids}\\
	    \cdashline{2-8}&&&&&&&&\\[-2.0ex]
        Model & 0.0 & 2.5 & 5.0 & 7.5 & 10.0 & 15.0 & 20.0 & \\
        \hline &&&&&&&&\\[-1.0ex]
        Na\"ive pricing & 	n.a. & 	n.a. & 	n.a. & 	n.a. & 	n.a. & 	n.a. & 	n.a. & 	n.a.	\\
        Logistic Regression & 	0.418 & 	0.465 & 	0.473 & 	0.475 & 	0.476 & 	0.476 & 	0.477 & 	0.447	\\
        Random Forest & 	0.411 & 	\textit{0.460} & 	\textit{0.469} & 	\textbf{0.472} & 	\textbf{0.473} & 	\textbf{0.473} & 	\textbf{0.474} & 	\textit{0.442}	\\
        MLP & 	\textbf{0.408} & 	\textbf{0.459} & 	\textbf{0.468} & 	\textbf{0.472} & 	\textbf{0.473} & 	\textbf{0.473} & 	\textbf{0.474} & 	\textbf{0.439}	\\
        \hdashline&&&&&&&&\\[-2.0ex]								
        HIE & 	0.413 & 	0.467 & 	0.476 & 	0.478 & 	0.479 & 	0.480 & 	0.480 & 	0.450	\\
        DRNets & 	\textit{0.410} & 	\textit{0.460} & 	0.470 & 	\textit{0.473} & 	\textbf{0.473} & 	\textit{0.474} & 	\textit{0.475} & 	\textit{0.442}	\\
        VCNets & 	0.415 & 	0.463 & 	0.471 & 	0.474 & 	\textit{0.475} & 	0.476 & 	0.476 & 	0.443	\\
        \hline \\[-1.0ex]
    \end{tabular}\\
    \footnotesize n.a.: not applicable
    \label{tbl:Results_BSs_SS}
\end{table}

\begin{table}[ht]
    \centering
    \caption{Mean Integrated Squared Error over expected revenue (MISE R) on Richards curve}
    \footnotesize
    \begin{tabular}{lccccccc:c}
        \multicolumn{9}{l}{\textbf{Ground truth:} Richards curve}\\
            \hline\\[-2.3ex]
	    \hline &&&&&&&&\\[-2.0ex] 
	    & \multicolumn{7}{c:}{\textit{Bias strength}} & \multirow{2}{*}{True bids}\\
	    \cdashline{2-8}&&&&&&&&\\[-2.0ex]
        Model & 0.0 & 2.5 & 5.0 & 7.5 & 10.0 & 15.0 & 20.0 & \\
        \hline &&&&&&&&\\[-1.0ex]
        Na\"ive pricing & 	n.a. & 	n.a. & 	n.a. & 	n.a. & 	n.a. & 	n.a. & 	n.a. & 	n.a.	\\
        Logistic Regression & 	0.063 & 	0.054 & 	0.055 & 	\textit{0.055} & 	\textit{0.058} & 	\textit{0.060} & 	\textit{0.065} & 	0.089	\\
        Random Forest & 	0.049 & 	0.069 & 	0.081 & 	0.093 & 	0.097 & 	0.106 & 	0.116 & 	0.125	\\
        MLP & 	\textbf{0.036} & 	\textbf{0.041} & 	\textbf{0.042} & 	\textbf{0.043} & 	\textbf{0.046} & 	\textbf{0.043} & 	\textbf{0.045} & 	\textbf{0.049}	\\
        \hdashline&&&&&&&&\\[-2.0ex]									
        HIE & 	0.063 & 	0.065 & 	0.066 & 	0.068 & 	0.067 & 	0.069 & 	0.071 & 	0.071	\\
        DRNets & 	\textit{0.041} & 	\textit{0.047} & 	0.074 & 	0.084 & 	0.060 & 	0.062 & 	0.088 & 	0.074	\\
        VCNets & 	0.048 & 	0.050 & 	\textit{0.053} & 	0.059 & 	0.065 & 	0.068 & 	0.073 & 	\textit{0.067}	\\
        \hline \\[-1.0ex]
    \end{tabular}\\
    \footnotesize n.a.: not applicable
    \label{tbl:Results_MISERs}
\end{table}

\begin{table}[!ht]
    \centering
    \caption{Mean Integrated Squared Error over expected revenue (MISE R) on stacked sigmoid}
    \footnotesize
    \begin{tabular}{lccccccc:c}
        \multicolumn{9}{l}{\textbf{Ground truth:} Stacked sigmoid}\\
            \hline\\[-2.3ex]
	    \hline &&&&&&&&\\[-2.0ex] 
	    & \multicolumn{7}{c:}{\textit{Bias strength}} & \multirow{2}{*}{True bids}\\
	    \cdashline{2-8}&&&&&&&&\\[-2.0ex]
        Model & 0.0 & 2.5 & 5.0 & 7.5 & 10.0 & 15.0 & 20.0 & \\
        \hline &&&&&&&&\\[-1.0ex]
        Na\"ive pricing & 	n.a. & 	n.a. & 	n.a. & 	n.a. & 	n.a. & 	n.a. & 	n.a. & 	n.a.	\\
        Logistic Regression & 	0.067 & 	0.080 & 	0.086 & 	0.091 & 	0.093 & 	\textit{0.098} & 	\textit{0.098} & 	\textit{0.074}	\\
        Random Forest & 	0.057 & 	0.096 & 	0.113 & 	0.129 & 	0.128 & 	0.139 & 	0.148 & 	0.165	\\
        MLP & 	\textbf{0.033} & 	\textit{0.049} & 	0.076 & 	\textit{0.081} & 	0.093 & 	0.104 & 	0.103 & 	0.109	\\
        \hdashline&&&&&&&&\\[-2.0ex]									
        HIE & 	0.048 & 	0.061 & 	\textit{0.070} & 	0.085 & 	\textit{0.084} & 	0.114 & 	0.122 & 	0.144	\\
        DRNets & 	\textit{0.043} & 	\textbf{0.046} & 	\textbf{0.050} & 	\textbf{0.055} & 	\textbf{0.053} & 	\textbf{0.070} & 	\textbf{0.064} & 	\textbf{0.067}	\\
        VCNets & 	0.056 & 	0.070 & 	0.090 & 	0.097 & 	0.096 & 	0.110 & 	0.109 & 	0.092	\\
        \hline \\[-1.0ex]
    \end{tabular}\\
    \footnotesize n.a.: not applicable
    \label{tbl:Results_MISERs_SS}
\end{table}

\pagebreak \null \newpage

\section{Hyperparameters\label{sec:Append_Hyperparam}}

\noindent All models trained for the experiments in this manuscript have been trained on a set of different hyperparameters, as described below.

\begin{table}[h]
    \centering
    \caption{Hyperparameters Random Forest}
    \footnotesize
    \centering
    \begin{tabular}[c]{lc}
        \hline 
        \hline \\[-2.0ex] 
        Parameter & Values \\
        \hline \\[-1.0ex]
        Number of trees & $\{ 100, 500\}$ \\
        Criterion & $\{ Gini\}$ \\
        Maximal depth per tree & $\{ None, 10\}$ \\[-0.1ex]
        \hline
    \end{tabular}
\end{table}

\begin{table}[h]
    \centering
    \caption{Hyperparameters MLP}
    \footnotesize
    \centering
    \begin{tabular}[c]{lc}
        \hline 
        \hline \\[-2.0ex] 
        Parameter & Values \\
        \hline \\[-1.0ex]
        Number of hidden layers & $\{ 2\}$ \\
        Nodes per hidden layer & $\{ 32, 48\}$ \\
        Batch size & $\{ 64, 128\}$ \\
        Number of steps & $\{ 1000, 2000\}$ \\
        Learning rate & $\{ 0.01, 0.05\}$ \\[-0.1ex]
        \hline
    \end{tabular}
\end{table}

\begin{table}[h]
    \centering
    \caption{Hyperparameters DRNet}
    \footnotesize
    \centering
    \begin{tabular}[c]{lc}
        \hline 
        \hline \\[-2.0ex] 
        Parameter & Values \\
        \hline \\[-1.0ex]
        Dosage strata & $\{ 10\}$ \\
        Number of representation layers  & $\{ 2\}$ \\
        Number of inference layers & $\{ 2\}$ \\
        Nodes per hidden layer & $\{ 32, 48\}$ \\
        Batch size & $\{ 64, 128\}$ \\
        Number of steps & $\{ 1000, 2000\}$ \\
        Learning rate & $\{ 0.01, 0.05\}$ \\[-0.1ex]
        \hline
    \end{tabular}
\end{table}
\end{appendix}
	
\end{document}